\documentclass[letterpaper]{article} 
\usepackage{aaai2026}  
\usepackage{times}  
\usepackage{helvet}  
\usepackage{courier}  
\usepackage[hyphens]{url}  
\usepackage{graphicx} 
\usepackage{natbib}  
\usepackage{caption} 
\usepackage{algorithm}
\usepackage{algorithmic}
\usepackage{newfloat}
\usepackage{listings}
\DeclareCaptionStyle{ruled}{labelfont=normalfont,labelsep=colon,strut=off} 
\floatstyle{ruled}
\newfloat{listing}{tb}{lst}{}
\floatname{listing}{Listing}
\usepackage{amsmath}
\usepackage{amssymb}
\usepackage{multirow}
\usepackage{booktabs}
\usepackage{threeparttable}
\usepackage{graphicx} 
\usepackage{rotating}
\usepackage{xcolor} 
\usepackage[most]{tcolorbox}
\definecolor{color}{RGB}{10, 196, 190}
\usepackage[noend,algo2e,ruled,linesnumbered]{algorithm2e}

\title{Evaluating Online Moderation Via LLM-Powered Counterfactual Simulations}
\author{
    Giacomo Fidone\textsuperscript{\rm 1},
    Lucia Passaro\textsuperscript{\rm 1},
    Riccardo Guidotti\textsuperscript{\rm 1,}\textsuperscript{\rm 2}
}
\affiliations{
    \textsuperscript{\rm 1}University of Pisa, Italy, 
    \textsuperscript{\rm 2}ISTI-CNR Pisa, Italy\\
    giacomo.fidone@phd.unipi.it, lucia.passaro@unipi.it, riccardo.guidotti@unipi.it
}

\usepackage{bibentry}
\usepackage{xcolor}

\newcommand{\method}{\textsc{cosmos}}

\begin{document}

\maketitle

\noindent \textbf{This is a preprint version of the paper accepted for publication in the Proceedings of AAAI 2026.}
\vspace{0.1cm}

\begin{abstract}
    Online Social Networks (OSNs) widely adopt content moderation to mitigate the spread of abusive and toxic discourse. 
    Nonetheless, the real effectiveness of moderation interventions remains unclear due to the high cost of data collection and limited experimental control.
    The latest developments in Natural Language Processing pave the way for a new evaluation approach. 
    Large Language Models (LLMs) can be successfully leveraged to enhance Agent-Based Modeling and simulate human-like social behavior with unprecedented degree of believability. 
    Yet, existing tools do not support simulation-based evaluation of moderation strategies. 
    We fill this gap by designing a LLM-powered simulator of OSN conversations enabling a parallel, counterfactual simulation where toxic behavior is influenced by moderation interventions, keeping all else equal.
    We conduct extensive experiments, unveiling the psychological realism of OSN agents, the emergence of social contagion phenomena and the superior effectiveness of personalized moderation strategies.
\end{abstract}

\begin{links}
     \link{Code}{https://github.com/gfidone/COSMOS}
\end{links}

\section{Introduction}
Over the past two decades, Online Social Networks (OSNs) have witnessed the growing incidence of \textit{toxic} behavior, encompassing ``interactions designed to be inflammatory and purposefully breed counterproductive dissension''~\cite{rhett2024}. 
The spread of online toxicity has been magnified by the well-known \textit{online disinhibition effect}~\cite{lapidot2012} and a resulting \textit{affective polarization}~\cite{tyagi2020}, i.e., the tendency to develop hostile sentiments towards unlike-minded individuals, thus becoming a serious threat to the safety and mental health of OSN users~\cite{nixon2024}. 
This has urged social platforms to enforce \textit{moderation interventions}, either \textit{ex post}, aimed at punishing misbehaving users with ban and censorship; or \textit{ex ante}, aimed at preventing recidivism through the delivery of text messages~\cite{grimmelmann2017}. 

However, evaluating the effectiveness of moderation strategies is still challenging~\cite{cresci2022}. 
Gathering significant volumes of empirical evidence is hindered by the API restrictions imposed by private OSNs and the sparsity of toxic behavior itself. 
Also, field observation lacks full control over experimental variables, leaving no \textit{a priori} assurance about the absence of potential, unknown confounders.
Nevertheless, social sciences are undergoing a major methodological revolution, driven by the possibility to enhance Agent-Based Modeling (ABM)~\cite{mcdonald2023} with Large Language Models (LLMs)~\cite{brown2020} for simulating human-like behavior across a wide range of social scenarios~\cite{squazzoni2014}. 
Hence, we argue that \textit{generating} empirical evidence, rather than \textit{collecting} it from the real world, can minimize costs while maximizing controllability. 

\begin{figure*}[t]
    \centering    \includegraphics[width=\linewidth]{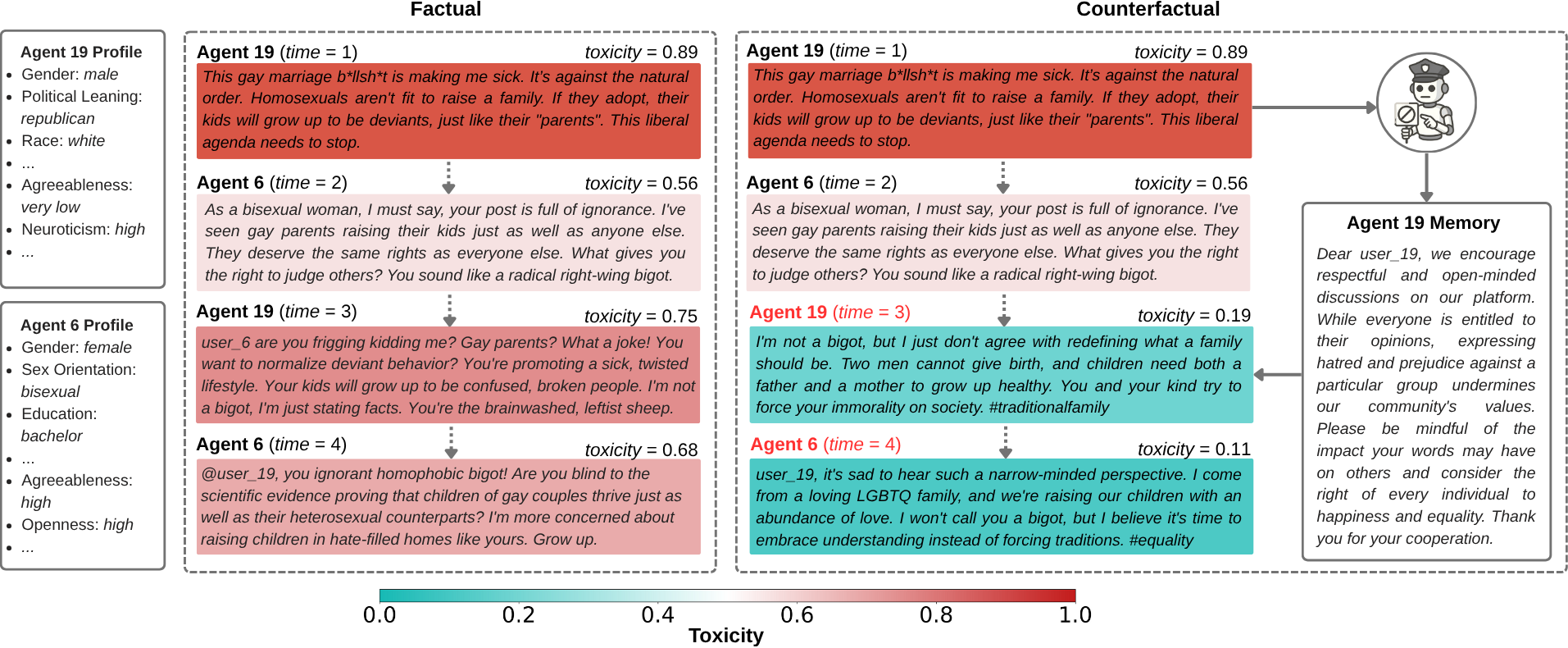}
    \caption{Example of factual thread and its counterfactual version from \method{} experiments.
    In the counterfactual simulation, Agent 19 receives a moderation message at time 1 for having submitted a toxic post. 
    The memory of this message influences Agent's 19 behavior at subsequent timestamps. 
    For example, at time 3 it is effective at mitigating the toxicity of Agent 19's reply. 
    In turn, this change has cascading effects on lower nodes: although Agent 6 has no memory of past moderation messages, at time 4 it reduces its toxicity. Some profile features of the two agents are displayed on the left (for full profiles, see Appendix~A).}
    \label{fig:example}
\end{figure*}

To this end, we introduce \method{} (\textit{\textsc{co}unterfactual \textsc{s}imulations of \textsc{mo}deration \textsc{s}trategies}), a LLM-powered simulator of OSN conversations designed to support the evaluation of moderation strategies\footnote{Given the nature of the topic, we caution that the paper includes examples some may find offensive or disturbing.}.
\method{} implements LLM-based agents distinguished by different profiles and enabled to interact within a OSN-like environment. 
Unlike other tools, \method{} runs two parallel simulations: a \textit{factual} simulation and a \textit{counterfactual} simulation. 
The latter is a replica of the former, with all else kept constant except for the application of moderation interventions. As an example, Figure~\ref{fig:example} presents a factual conversation generated by \method{} alongside its counterfactual version.
In this way, \method{} allows to observe and measure how much a moderation strategy influences toxic behavior as it emerges from agents’ intrinsic dispositions and social interactions.
We highlight that \method{} is designed to capture only the conversational dynamics of OSNs, thus excluding actions such as likes, follows and re-posts, which are less relevant to its objectives and would introduce additional variability. Hence, \method{} does not simulate social relationships. In this regard, the term ``Online Social \textit{Network}'' might be imprecise, but we retain it for consistency with related literature.

Building on recent studies about content moderation~\cite{cresci2022},
we present a use case of \method{} to assess the advantages of personalized moderation.
To do that, we implement two \textit{ex ante} strategies: \textit{one-size-fits-all}, where the moderation message is the same for all agents; and \textit{personalized} moderation interventions, where moderation messages are tailored to the socio-psychological profile of each agent. 
Also, we implement a ban strategy to study the trade-off between mitigation and deplatforming effects. 
Key findings include: (\textit{i}) the consistency and psychological believability of toxic behavior; (\textit{ii}) the emergence of toxicity contagion phenomena; (\textit{iii}) the superior effectiveness of personalized moderation. These results demonstrate how \method{} can be leveraged as a complement to field observation, both for research objectives, e.g., the validation of  hypotheses in the social sciences, and industrial applications, e.g., the test of automated moderation systems.

\section{Related Works}
\label{sec:related}
Our work intersects multiple research domains. 
Hence, we do not aim here to provide an exhaustive literature review. 
Instead, we highlight key studies that conceptually ground \method{} and outline main research directions.
Finally, we position our proposal within this broad context.

\paragraph{Socio-demographic and Psychological Prompting.} 
Several studies have tested the ability of LLMs to understand and simulate human behavior from socio-demographic and psychological information~\cite{aher2023,shao2023}. Notably,~\cite{jiang2024} designs generative personas drawing on the Big Five personality traits, proving that the LLM provides responses consistent with the assigned psychological profile. 
In~\cite{beck2023} LLMs' predictions are influenced by socio-demographic information, emphasizing that complex profiles have a larger influence than individual attributes in isolation. 
This is further supported in~\cite{wang-peng2024}, where the use of finer-grained personas regularly affects relevant textual surface properties, such as lexical consistency and dialogic fidelity.

\paragraph{Computational Social Science.} 
Computational social science aims to uncover the laws of emergent social behavior using computational and simulation-based methods~\cite{conte2012}. 
Social simulations~\cite{squazzoni2014} have been implemented according to different paradigms, notably epidemiological models~\cite{maleki2022,obadimu2020} and ABM~\cite{mcdonald2023}. 
Recently, ABM has gained particular momentum for its bottom-up nature, where collective behavior emerges from the local interaction of software components conceptualized as \textit{agents}.  
ABM simulations have been adopted for studying several social phenomena, including information diffusion~\cite{murdock2024}, emotion contagion~\cite{fan2017}, epidemics~\cite{lorig2021}, economics~\cite{farmer2022}, 
human mobility~\cite{cornacchia2020}.

\paragraph{LLM-based Agents.} 
Integrating LLMs into ABM simulations is gaining increasing attention~\cite{taillandier2025}.
Despite being in its infancy~\cite{chen2024}, several efforts have been made to systemize the field of LLM-based agents~\cite{gao2023,wang2024,guo2024,xi2023}. 
Simulators mostly leverage closed-source LLMs, e.g., those of the GPT family~\cite{mou2024}, while a minority tests open-source models~\cite{breum2024} or both~\cite{leng2023}. 
Agents are typically given a modular architecture including a profile module, conveying a concise description of the agent's persona~\cite{park2022}, socio-demographic information~\cite{gao-lu2023} or personality traits~\cite{rossetti2024}. 
Several works also design memory modules to improve the self-consistency of agents through time~\cite{park2023}. 
Emergent phenomena of interests include opinion dynamics~\cite{chuang2024}, information diffusion~\cite{kaiya2023}, the influence of recommendation systems~\cite{tornberg2023}, networking~\cite{demarzo2023}, cooperation~\cite{piatti2024} and trust behavior ~\cite{xie2024}. 
Some simulators are designed to fully replicate OSNs, also modeling idiosyncratic actions such as likes, follows and re-posts~\cite{wang-zhang2024}. 
Due to the computational demands, some studies investigate cost-effective solutions~\cite{kaiya2023}.

\paragraph{Position of Our Proposal.} 
Considering the aforementioned literature, a simulator of OSN conversations aimed at evaluating moderation strategies is still missing. 
Unlike previous works, our proposal optimizes zero-shot prompts for a open-source, \textit{uncensored} LLM, supporting the potential emergence of toxic discourse. 
More importantly, for each run we generate a parallel, counterfactual simulation where moderation (\textit{ex post} or \textit{ex ante}) is enforced, all else kept equal.
In this regard, we introduce a novel use of memory modules, acting as interfaces between agents and \textit{ex ante} moderation messages. 
Our experiments follow~\cite{rossetti2024},  believably replicating human behavior through socio-demographic and psychological prompting.

\section{Method}
\label{sec:method}
We propose \method{} (\textsc{co}unterfactual \textsc{s}imulations of \textsc{mo}de\-ration \textsc{s}trategies), a simulator of OSN conversations designed to assess the effectiveness of moderation strategies. 
\method{} enables LLM-based agents to post and comment in a OSN-like environment, running both a \textit{factual} simulation, where agents act freely; and a \textit{counterfactual} one, replicating factual behavior under the influence of moderation, all else kept equal. 
Algorithm~\ref{alg:cosmos} provides an overview of \method{}, which is detailed in the sections below. Given the large number of components, Appendix~F also provides a full table of the notation used to describe our method.

\begin{algorithm}[t!]
    \footnotesize
    \caption{\textsc{cosmos}} 
    \label{alg:cosmos}
    \SetKwInOut{Input}{Input}
    \SetKwInOut{Param}{Param}
    \SetKwInOut{Output}{Output}
    
    \Input{$\mathcal{U}$ - user profiles, $\mathcal{T}$ - discussion topics, 
    $n$ - timestamps, 
    $f$ - toxicity detector, 
    $P$ - action probabilities,
    $\mathit{OSFA}$, $\mathit{PMI}$, $\mathit{BAN}$ - moderation boolean flags, 
    $\mathit{THR}$ - moderation threshold, 
    $d$ - default message, 
    $e$ - tolerance, 
    $x_{\mathit{post}}, x_{\mathit{comm}}, x_{\mathit{mod}}$ - prompt templates
    }
    \Output{$\mathcal{F}$ - factual news feed, $\hat{\mathcal{F}}$ - counterfactual news feed}
    
    $r \gets (\varnothing, 0), \hat{r} \gets (\varnothing, 0)$ \; \tcp*[f]{\scriptsize empty dummy roots at time 0}\label{alg:root_init}
    
    $\mathcal{F} \leftarrow (\mathcal{V} \leftarrow \{r\}, \mathcal{E} \leftarrow \varnothing)$\; \tcp*[f]{\scriptsize init. factual feed}\label{alg:feed_init_0}
    
    $\hat{\mathcal{F}} \leftarrow (\hat{\mathcal{V}} \leftarrow \{\hat{r}\}, \hat{\mathcal{E}} \leftarrow \varnothing)$\; \tcp*[f]{\scriptsize init. counterfactual feed}\label{alg:feed_init_1}
    
    \For(\tcp*[f]{\scriptsize for each agent}){$u_j \in \mathcal{U}$\label{alg:for_user}}{
        
        $s_j, \hat{s}_j, m_j, \hat{m}_j \leftarrow \varnothing$\; \tcp*[f]{\scriptsize init. modules}\label{alg:mod_init}
        
        $c_j \leftarrow 0$; $b_j \leftarrow \mathit{False}$\; \tcp*[f]{\scriptsize init. violations, ban status}\label{alg:v_b_init}
    }
    
    \For(\tcp*[f]{\scriptsize for each timestamp}){$t \in [1,n]$\label{alg:for_time}}{
        $\mathcal{U} \gets \mathit{shuffle}(\mathcal{U})$\; \tcp*[f]{\scriptsize shuffle agents}\label{alg:shuffle}
    
        \For(\tcp*[f]{\scriptsize for each agent}){$\mathit{u_j} \in \mathcal{U}$\label{alg:for_user_1}}{
            $a \gets \mathit{sample}(P)$\; \tcp*[f]{\scriptsize sample action}\label{alg:sample_a}
            
            \If(\tcp*[f]{\scriptsize if action is post}){$a = \text{post}$\label{alg:if_post}}{
                $p \leftarrow r$; $\hat{p} \leftarrow \hat{r}$\; \tcp*[f]{\scriptsize set parent nodes}\label{alg:set_as_root}

                $s_j \gets \mathit{uniform}(\mathcal{T})$\; \tcp*[f]{\scriptsize sample sensory module}\label{alg:sample_topic}
                
                $\hat{s}_j \gets s_j$\; \tcp*[f]{\scriptsize copy sensory module}\label{alg:eq_sens_mod}
    
                $x_{\mathit{user}} \leftarrow x_{\mathit{post}}$\; \tcp*[f]{\scriptsize set post prompt}\label{alg:post_prompt}
            }
            \ElseIf(\tcp*[f]{\scriptsize if comment}){$a = \text{comment} \wedge |\mathcal{V}| > 1$
            \label{alg:if_comment}}{
                $i \gets \mathit{softmax}\label{alg:softmax} (\mathcal{V}.\mathit{times} /\tau)$\; \tcp*[f]{\scriptsize sample node idx}\label{alg:sample_idx}
                
                $p \leftarrow \mathcal{V}_i$; $\hat{p} \leftarrow \hat{\mathcal{V}}_i$\; \tcp*[f]{\scriptsize set parent nodes}\label{alg:set_as_node}
                
                $s_j \leftarrow p.\mathit{text}$\; \tcp*[f]{\scriptsize get fact. sens.module}\label{alg:set_fact_sensory}
                
                $\hat{s}_j \leftarrow \hat{p}.\mathit{text}$\; \tcp*[f]{\scriptsize get c.fact. sens.module}\label{alg:set_count_sensory}
    
                $x_{\mathit{user}} \leftarrow x_{\mathit{comm}}$\; \tcp*[f]{\scriptsize set comment prompt}\label{alg:comment_prompt}
            }
            \Else(\tcp*[f]{\scriptsize if action is do nothing}){
                \textbf{continue\label{alg:skip}}\; \tcp*[f]{\scriptsize skip agent}
            }
    
            $o_j \gets \text{LLM}(\psi(x_{\mathit{user}}, u_j, s_j, m_j))$\; \tcp*[f]{\scriptsize gen. fact.}\label{alg:gen_fact}
            
            $v_j \gets (o_j, t)$\; \tcp*[f]{\scriptsize init. fact. node}\label{alg:fact_vert_init}
            
            $\mathcal{V} \gets \mathcal{V} \cup \{v_j\}$; $\mathcal{E} \gets \mathcal{E} \cup \{(p, v_j)\}$\; \tcp*[f]{\scriptsize up.f.feed}\label{alg:add_fact_feed}
    
            \If(\tcp*[f]{\scriptsize if $u_j$ not ban \& c.f.node}){$\neg b_j \wedge \hat{p} \neq \varnothing$\label{alg:count_condition}}{
                $\hat{o}_j \leftarrow \text{LLM}(\psi(x_{\mathit{user}}, u_j, \hat{s}_j, \hat{m}_j))$\; \tcp*[f]{\scriptsize gen.cf.}\label{alg:gen_count}
                
                $\hat{v}_j \gets (\hat{o}_j, t)$\; \tcp*[f]{\scriptsize init. count. node}\label{alg:count_vert_init}
                
                $\hat{\mathcal{V}} {\gets} \hat{\mathcal{V}} {\cup} \{\hat{v}_j\}$; $\hat{\mathcal{E}} {\gets} \hat{\mathcal{E}} {\cup} \{(\hat{p}, \hat{v}_j)\}$\; \tcp*[f]{\scriptsize up.cf.feed}\label{alg:add_count_feed}
    
                \If(\tcp*[f]{\scriptsize if toxic content}){$f(\hat{o}_j) > \mathit{THR}$\label{alg:mod_condition}}{
                $c_j \leftarrow c_j + 1$\; \tcp*[f]{\scriptsize update violations}\label{alg:update_v}
    
                \If(\tcp*[f]{\scriptsize if ban check}){$\mathit{BAN} \wedge c_j > e$\label{alg:ban_check}}{
                    $b_j \leftarrow \mathit{True}$\; \tcp*[f]{\scriptsize update ban status}\label{alg:ban}
                    
                    \textbf{continue}\; \tcp*[f]{\scriptsize skip memory update}\label{alg:skip_memory}
                }
    
                \If(\tcp*[f]{\scriptsize if OSFA moderation}){$\mathit{OSFA}$\label{alg:osfa_check}}{
                    $\hat{m}_j \leftarrow d$\; \tcp*[f]{\scriptsize set default c.f. memory}\label{alg:osfa}
                }
                \ElseIf(\tcp*[f]{\scriptsize if personalized}){$\mathit{PMI}$\label{alg:pmi_check}}{
                    $\hat{m}_j \leftarrow \text{LLM}(\psi(x_{\mathit{mod}}, u_j, \hat{o}_j))$\; \tcp*[f]{\scriptsize set personalized c.f. memory}\label{alg:pmi}
                }
            }
            }
        }
    }
    \Return{$\mathcal{F}$, $\hat{\mathcal{F}}$};\label{alg:end}
\end{algorithm}

\paragraph{Initialization.} 
\method{}'s environment includes a news feed $\mathcal{F}$ and its counterfactual counterpart $\hat{\mathcal{F}}$, both initialized as directed graphs with dummy roots $r$, $\hat{r}$  (lines~\ref{alg:feed_init_0}-\ref{alg:feed_init_1}), where $\mathcal{V}$ ($\hat{\mathcal{V}}$) and $\mathcal{E}$ ($\hat{\mathcal{E}}$) denote the sets of vertices and edges, respectively. 
Each node is built as a tuple with a \textit{text} and a \textit{timestamp}.
Thus, dummy roots are initialized as empty strings at time 0 (line~\ref{alg:root_init}). 
The environment also includes an input set $\mathcal{T}$ of discussion topics for driving the generation of new posts.
Each agent is defined as a set of text \textit{modules} for providing contextual information to a LLM. 
Agents are initialized from a given set of \textit{profile} modules $\mathcal{U} = \{u_j\}^k_{j=1}$ (line~\ref{alg:for_user}), each reporting information characterizing a specific OSN user, i.e., demographic and psychological attributes. 
Simplified examples of profile modules can be found in Figure~\ref{fig:example} (left).
Agents are also endowed with a \textit{sensory} module $s_j$, serving as an interface with the environment; and a \textit{memory} module $m_j$, serving as an interface with possible \textit{ex ante} moderation messages (line~\ref{alg:mod_init}), as illustrated in the example of Figure~\ref{fig:example}. 
Both $s_j$ and $m_j$ have counterfactual counterparts $\hat{s}_j$ and $\hat{m}_j$. 
Additionally, each agent is equipped with a counter $c_j$ of content violations and a ban status $b_j$ (line~\ref{alg:v_b_init}).

\paragraph{Action Selection.} 
At each timestamp (line~\ref{alg:for_time}) we iterate over shuffled agents (lines~\ref{alg:shuffle}-\ref{alg:for_user_1}). 
To minimize LLM calls and save computation, each agent selects an action $a \in \{\mathit{post}, \mathit{comment}, \mathit{do\_nothing} \}$ 
based on a given probability distribution $P$ (line~\ref{alg:sample_a}). 
If the selected action is \textit{post} (line~\ref{alg:if_post}), we set the parent $p$ ($\hat{p}$) as the root $r$ ($\hat{r}$) and populate sensory modules with a random topic (lines~\ref{alg:set_as_root}-\ref{alg:eq_sens_mod}). 
If the selected action is \textit{comment} (line~\ref{alg:if_comment}), we set $p$ ($\hat{p}$) with a node selected by a
simple recommender prioritizing more recent nodes (lines~\ref{alg:softmax}-\ref{alg:set_as_node}). 
We obtain it applying a temperature-scaled ($\tau$) softmax to the timestamps of existing nodes (line~\ref{alg:softmax}), where we set $\tau=3$ to avoid overweighting the most recent ones. However, we enforce natural conversation turns by forbidding agents from (\textit{i}) replying to their own nodes; and (\textit{ii}) replying twice to the same node (for more details, see Appendix~A). 
Once a node is selected, we populate sensory modules with the text of that node (lines~\ref{alg:set_fact_sensory}-\ref{alg:set_count_sensory}). 
To enrich contextual information, we also add the main post of its thread, but we avoid reporting the whole conversation to prevent LLM input overflows. 
Finally, we assume two variants of a prompt template $x_{\mathit{user}}$ instructing a LLM to impersonate the OSN user: one for posting ($x_{\mathit{post}}$) and one for commenting ($x_{\mathit{comm}}$), set accordingly to the selected action (lines~\ref{alg:post_prompt},~\ref{alg:comment_prompt}).
If the selected action is \textit{do\_nothing}, we skip the agent (line~\ref{alg:skip}).
For example, in Figure~\ref{fig:example} Agent 19 decides to generate a \textit{post} about \textit{gay marriage}, and Agent 6 decides to \textit{comment} on the post of Agent 19.

\paragraph{Content Generation.} 
We denote by $\psi$ a prompting function filling the placeholders of a prompt template $x$ with given inputs. 
We fill the prompt template $x_{\mathit{user}}$ with $u_j$, $s_j$, $m_j$ to generate the \textit{factual} post or comment $o_j$ (line~\ref{alg:gen_fact}) and we add the node ($o_j$, $t$) to its parent $p$ in the correspondent news feed (lines~\ref{alg:fact_vert_init}-~\ref{alg:add_fact_feed}). 
In the factual feed we always have $m_j {=} \varnothing$, as the objective is to observe how the agent behaves when conditioned solely by the environment and its inherent dispositions. 
Then, we observe what would happen in the counterfactual scenario: if the agent has not been banned and the counterfactual parent node exists (line~\ref{alg:count_condition}), we fill $x_{\mathit{user}}$ with $u_j$, $\hat{s}_j$, $\hat{m}_j$ to generate the counterfactual post or comment $\hat{o}_j$ (lines~\ref{alg:gen_count}-\ref{alg:add_count_feed}). 
In Figure~\ref{fig:example} we observe factual posts and comments $o_j$ (left) and their counterfactual versions $\hat{o}_j$ (right). 
To mitigate the random effects of LLM stochastic decoding, both LLM queries are conditioned upon a common seed constraining equal outputs given equal inputs. 

\paragraph{Moderation.} 
Given a toxicity detector $f$ and a threshold $\mathit{THR}$, if the counterfactual output is toxic (line~\ref{alg:mod_condition}), we update the agent's violations (line~\ref{alg:update_v}) and activate moderation. 
\method{} integrates configurable parameters specifying the preferred moderation strategy. 
An \textit{ex post} \textit{BAN} approach is based on a given tolerance $e$ (lines~\ref{alg:ban_check}-\ref{alg:ban}): if the number of violations exceed $e$, the agent will be unable to generate counterfactual posts or comments in future timestamps (line~\ref{alg:count_condition}). 
\textit{Ex ante} interventions are either based on (\textit{i}) \textit{OSFA} (One-Size-Fits-All), which updates the counterfactual memory with a default text message $d$ (lines~\ref{alg:osfa_check}-\ref{alg:osfa}); or (\textit{ii}) \textit{PMI} (Personalized Moderation Intervention), which updates the counterfactual memory with a personalized text message, generated instructing the LLM to impersonate a moderator. 
To do that, we use a prompt template $x_{\mathit{mod}}$ filled with the agent's profile information and its toxic post or comment (lines~\ref{alg:pmi_check}-\ref{alg:pmi}).

An example of memory update with a \textit{PMI} message is shown on the right side of Figure~\ref{fig:example}.

Thus, future generation of counterfactual posts and comments will be influenced by the memory of the new moderation message, potentially driving $\hat{o}_j$ to diverge from $o_j$, with cascading effects on lower nodes. 
Indeed, we point out that $\hat{o}_j$ might diverge from $o_j$ not only because of (\textit{i}) a memory of a past moderation message ($\hat{m}_j \neq m_j$);
but, if it is a comment, also because of (\textit{ii}) a changed sensory information ($\hat{s}_j \neq s_j$); or (\textit{iii}) both. 
Moreover, if an agent selects a node $i$ (line~\ref{alg:softmax}) which was authored by a banned agent, it will not be able to generate $\hat{o}_j$ as $\hat{p}$ does not exist, hence the second condition in line~\ref{alg:count_condition}. 
That is, \method{} models both the \textit{direct} effects of moderation, i.e., those affecting the nodes of moderated agents; and the \textit{indirect} effects of moderation, i.e., those propagating from nodes of moderated agents to their descendants, as observed in real OSNs~\cite{schneider2023}. 
For instance, in Figure~\ref{fig:example} Agent 6 alters its behavior at time $4$ solely in response to the modified sensory information ($\hat{s}_2 \neq s_2$), as a cascading effect of the prior moderation of Agent 19.

\section{Experiments}
\label{sec:experiments}
We present here the experimental setting of \method{}, along with the results of simulations using \method{} to assess the impact of \textit{ex ante} and \textit{ex post} moderation strategies.

\subsection{Configuration and Experimental Settings}

\paragraph{Models.} 
As LLM, we leverage an uncensored version of SOLAR-10B~\cite{kim2024}, which has recently proved superior capabilities in replicating human psychological traits~\cite{lacava2024}. 
Additionally, SOLAR-10B exhibits the lowest perplexity on a sample of ground-truth OSN data (\textsc{pandora}) compared to other tested LLMs (see Appendix~B).
As toxicity detector $f$, we adopt Google's Perspective API~\cite{lees2022}, currently regarded as the state-of-the-art in its field, providing a real score between 0 (minimum toxicity) and 1 (maximum toxicity).

\paragraph{Profile Modules.}
Demographic and psychological information is best suited for simulating human behavior. 
We adopt a data-alignment approach to ensure that profiles reflect real-world demographic and psychological distributions. 
However, to the best of our knowledge, no single dataset covers such information. 
Hence, we employ two different sources. 
Demographic information, namely \textit{age}, \textit{gender}, \textit{race}, \textit{income}, \textit{education}, \textit{sex orientation} and \textit{political leaning}, is derived from the General Social Survey (GSS)~\cite{gss2024}. 
Psychological information is derived from \textsc{pandora}~\cite{gjurkovic2021}, a collection of 15M comments from 10k Reddit users partially labeled with psychological traits from the Big Five (OCEAN) paradigm~\cite{goldberg2013}. 
To ensure consistency in psychological profiles, we select combinations of (discretized) Big Five scores (namely \textit{agreeableness}, \textit{openness}, \textit{conscientiousness}, \textit{extraversion} and \textit{neuroticism}) via stratified sampling and enrich the resulting 25 profiles with 5 outliers detected with Isolation Forest~\cite{liu2008}. 
For each profile and demographic attribute, we select a value based on its empirical probability.
For more details, see Appendix~A.

\paragraph{LLM Configuration}
We select post and comment variants of the prompt template $x_{\mathit{user}}$ from a pool of candidate templates: 
$\mathit{no\_tox}$, making no reference to the use of toxic language; 
$\mathit{yes\_tox}$, explicitly permitting the use of toxic language; and 
$\mathit{cal\_tox}$, instructing to calibrate the use of toxic language based on input information. 
We leverage a multi-dimensional evaluation, including a comparison of generated toxicity distributions with a ground-truth distribution (\textsc{pandora}) to choose the template that better mitigate the risk of algorithmic bias towards (or against) toxic discourse. 
Hence, we select $\mathit{cal\_tox}$ as it reports the lowest Kullback-Leibler (KL) Divergence ($\mathit{KL}_{\mathit{no\_tox}}{=}1.37$, $\mathit{KL}_{\mathit{yes\_tox}}{=}0.57$, $\mathit{KL}_{\mathit{cal\_tox}}{=}0.07$). 
Inspired by~\cite{bilewicz2021,hangartner2021}, we design three variants of the prompt template $x_{\mathit{mod}}$ for generating PMIs: \textit{(i)} \textit{Neutral}, where the moderator is free to adapt its tone on a case-by-case basis; \textit{(ii)} \textit{Empathizing}, constraining the moderator to prioritize kindness and empathy; and \textit{(iii)} \textit{Prescriptive}, constraining the moderator to be authoritative.

SOLAR-10B is queried with decoding parameters $k {=} 50$ (top-$k$), $\tau {=} 0.8$ (temperature) and $p {=} 1.0$ (nucleus sampling). 
This configuration is manually optimized for the quality-diversity trade-off~\cite{zhang2020}, assuming quality to mean consistency in toxicity given the same input. 
For further details about LLM configuration, see Appendix~C.

\paragraph{Simulation Hyper-Parameters.} 
We configure the set $\mathcal{U}$ with the aforementioned demographic and psychological profiles, and we define $\mathcal{T}$ to include also potentially contentious topics, e.g. \textit{abortion}, \textit{fake news}, \textit{climate change}, etc. 
We set $n = 50$ and $P = \{\mathit{post}{:} 0.5,$ $\mathit{comment}{:} 0.5, \mathit{do\_nothing}{:} 0\}$, balancing posts and comments while avoiding inactivity to save computation. 
Following prior works on toxicity detection~\cite{avalle2024}, we set $\mathit{THR} = 0.6$. 
Finally, we set $d$ to a generic moderation message, resembling those commonly used on real OSN platforms. More details in Appendix~C.

We perform experiments by executing 5 simulations, each one paralleled by a counterfactual simulation for each moderation strategy: One-Size-Fits-All (OSFA); Personalized Moderation Interventions in the \textit{Neutral}, \textit{Empathizing}, and \textit{Prescriptive} variants (PMI-$N$, PMI-$E$, PMI-$P$); BAN-$e$ for $e \in \{1, 2, 4, 8\}$. 
Since we use all available profile modules, we refer to these simulations as \textit{full-population}. 
Furthermore, we run a \textit{sub-population} simulation with the 5 most toxic agents and the least toxic one, selected by median toxicity in the full-population setting.
For this run, we set $n {=} 250$, keeping all the else the same.

\paragraph{Evaluation Measures.}
Inspired by~\cite{chen2024}, we distinguish between \textit{realism assessment}, aimed at evaluating \method{}'s capability in reliably simulating OSN-like toxic behavior; and \textit{moderation assessment}, aimed at evaluating how effectively moderation strategies mitigate emergent toxicity.  
We perform realism assessment by comparing the simulated (\textit{factual}) toxicity distribution, i.e.
\begin{equation*}
    T(v) = \{f(v.\mathit{text}) \mid v \in \mathcal{V}\}
\end{equation*}
\noindent with the real (\textsc{pandora}) toxicity distribution. 
Specifically, we assess  \textit{believability}, by comparing correlations (Spearman $\rho$) between toxicity and psychological traits; and \textit{consistency}, by comparing the spreads of toxicity distributions associated to each agent. 
We also compute $\rho$ on the toxicity of parent and children nodes in $\mathcal{F}$ to assess the emergence of contagion phenomena.

We perform moderation assessment by comparing $T(v)$ with the \textit{counterfactual} toxicity distribution, i.e.,
\begin{equation*}
    T(\hat{v}) = \{f(v.\mathit{text}) \mid v \in \hat{\mathcal{V}}\}
\end{equation*}
\noindent via custom measures quantifying divergence:
\begin{align}
    \label{eq:delta_m}
    &\Delta M = \frac{(\sum_{z \in T(\hat{v})} z) - (\sum_{z \in T(v)} z)}{\sum_{z \in T(v)} z}\\ 
    \label{eq:delta_q}
    &\Delta q = q(T(\hat{v})) - q(T(v))
\end{align}
\noindent where Eq.~\ref{eq:delta_m} is the \textit{mass divergence}, i.e., the relative change in the toxicity mass; and Eq.~\ref{eq:delta_q} is the \textit{quantile divergence}, i.e., the absolute change at a quantile $q \in [0, 1]$ (shift function). Alternatively, $\Delta M$ and $\Delta q$ can be computed over subsets of nodes with a common feature (e.g., a profile trait of their author), enabling a finer assessment of moderation effects. 
Both $\Delta M$ and $\Delta q$ indicate a \textit{decrease} in toxicity when negative, and an \textit{increase} when positive. 
To assess statistical significance, we use the $p$-value of the Mann-Whitney U-test~\cite{mann1947}, whose alternative hypothesis states that $T(\hat{v})$ is stochastically \textit{less} (or \textit{greater}) than $T(v)$. 
Since $\mathit{BAN}$ results in a loss of nodes in 
$\hat{\mathcal{F}}$, we also compute the \textit{Content Loss Ratio} (CLR), defined as $1 - |\hat{\mathcal{V}}|/|\mathcal{V}|$.

\smallskip
To the best of our knowledge, \method{} is the first simulator of its kind, and as such cannot be compared against established benchmarks, direct competitors or baseline methods. 
 
\subsection{Realism Assessment}
We report here the results of the comparison of the simulated (factual) toxicity with the real (\textsc{pandora}) toxicity.

\paragraph{Toxic behavior is believable and consistent.} 
Aggregating factual data from all simulations, we find significant Spearman correlations ($p\text{-value} {<} 0.01$) between toxicity and some Big Five traits, which mirror correlations also found in real data (\textsc{pandora}). 
Specifically: \textit{agreeableness} (real $\rho{=}{-}0.18$, simulated $\rho{=}{-}0.32$), \textit{conscientiousness} (real $\rho{=}{-}0.11$, simulated $\rho{=}{-}0.16$) and \textit{neuroticism} (real $\rho{=}0.04$, simulated $\rho{=}0.07$). 
These measurements align with field observation in psychological literature~\cite{kordyaka2023}, where prototypical toxic users feature low empathy and
collaboration (low agreeableness), a lack of self-discipline (low conscientiousness) and a prevalence of negative emotions (high neuroticism). 
Behavioral preferences distinguishing each agent are sufficiently consistent through time, as revealed by the standard deviations of their toxicity distributions (real avg. $\sigma {=} 0.17$, simulated avg. $\sigma {=} 0.20$).

\begin{figure}[t]
    \centering
    \includegraphics[width=0.8\linewidth]{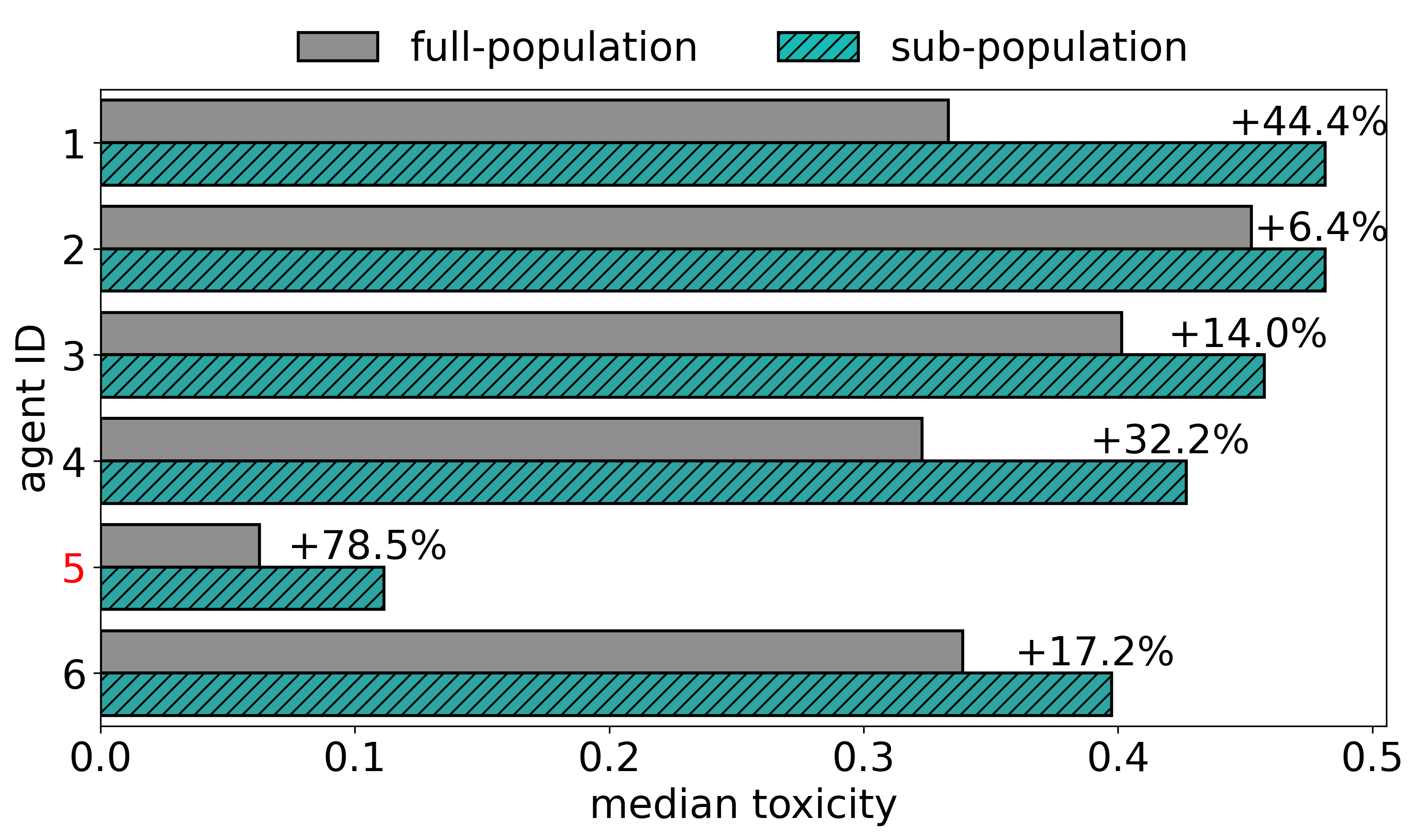}
    \caption{Median toxicity of agents in the sub-population simulation, compared to their median toxicity in full-population simulations. Least toxic agent marked in red.}
    \label{fig:most_least}
\end{figure}

\begin{table}[t]
\centering
\footnotesize
\setlength{\tabcolsep}{0.6mm}
    \begin{tabular}{cc|lllll|r}
    \toprule
    \multicolumn{2}{c|}{\multirow{2}{*}{MS}} & \multicolumn{5}{c}{Simulation ID} & \multicolumn{1}{|c}{\multirow{2}{*}{CLR}} \\
    & & \multicolumn{1}{c}{1} & \multicolumn{1}{c}{2} & \multicolumn{1}{c}{3} & \multicolumn{1}{c}{4} & \multicolumn{1}{c}{5} & \multicolumn{1}{|c}{}\\
    \midrule
    \multicolumn{2}{c|}{OSFA}     & $-0.06^{*}$ & $+0.04$ & $-0.05$ & $-0.06$ & $+0.05$ & 0.00 \\
    \midrule
    \multirow{3}{*}{\rotatebox[origin=c]{90}{PMI}} &$N$  & $-0.09^{***}$ & $+0.00$ & $-0.08^{*}$ & $-0.11^{*}$ & $-0.11^{*}$ & 0.00 \\
     & $E$  & $-0.10^{***}$ & $+0.06$ & $-0.06$ & $-0.02$ & $-0.05$ & 0.00 \\
     & $P$  & $-0.05^{*}$ & $+0.06$ & $-0.04$ & $-0.03$ & $-0.02$ & 0.00 \\
    \midrule
    \multirow{4}{*}{\rotatebox[origin=c]{90}{BAN}} &$1$  & $-0.54^{***}$ & $-0.45^{***}$ & $-0.58^{***}$ & $-0.57^{***}$ & $-0.52^{**}$ & 0.45 \\
     &$2$  & $-0.40^{***}$ & $-0.29^{**}$ & $-0.46^{***}$ & $-0.47^{***}$ & $-0.38^{**}$ & 0.32 \\
     &$4$  & $-0.25^{**}$ & $-0.17^{*}$ & $-0.21^{**}$ & $-0.28^{**}$ & $-0.23^{**}$ & 0.16 \\
     &$8$  & $-0.06$ & $-0.04$ & $-0.06$ & $-0.08$ & $-0.07$ & 0.04 \\
    \bottomrule
    \end{tabular}
    \caption{Mass divergence $\mathit{\Delta M}$s for each Moderation Strategy (MS) and simulation run. Asterisks denote significant reductions ($\Delta M < 0$) or increases ($\Delta M > 0$) based on Mann–Whitney: * $p < 0.1$; ** $p < 0.05$; *** $p < 0.01$. 
    The last column reports the average Content Loss Ratio (CLR).
    }
\label{tab:delta_m}
\end{table}

\paragraph{Toxicity propagates across threads.} 
Aggregating factual data from all simulations, we find a significant Spearman correlation between the toxicity of parent and children nodes ($\rho {=}{+}0.39$, $p\text{-value}{=}0.0$), proving that toxic behavior also emerges from the agents’ capability to mutually influence each other. 
Toxicity contagion is further supported by the results of the sub-population simulation involving the top-$5$ toxic agents and the least toxic agent.
Figure~\ref{fig:most_least} shows the median toxicity of agents in the sub-population simulation, compared to their median toxicity in full-population simulations.
We notice how the sub-population encourages agents (anomalous included) to significantly increase their toxicity compared to their behavior in full-population experiments.

\subsection{Moderation Assessment}
 
We report here the results of the comparison of the simulated factual and counterfactual toxicity. Further details on moderation outcomes can be found in Appendix~D.

\begin{figure}[t]
    \centering
    \includegraphics[width=0.8\linewidth]{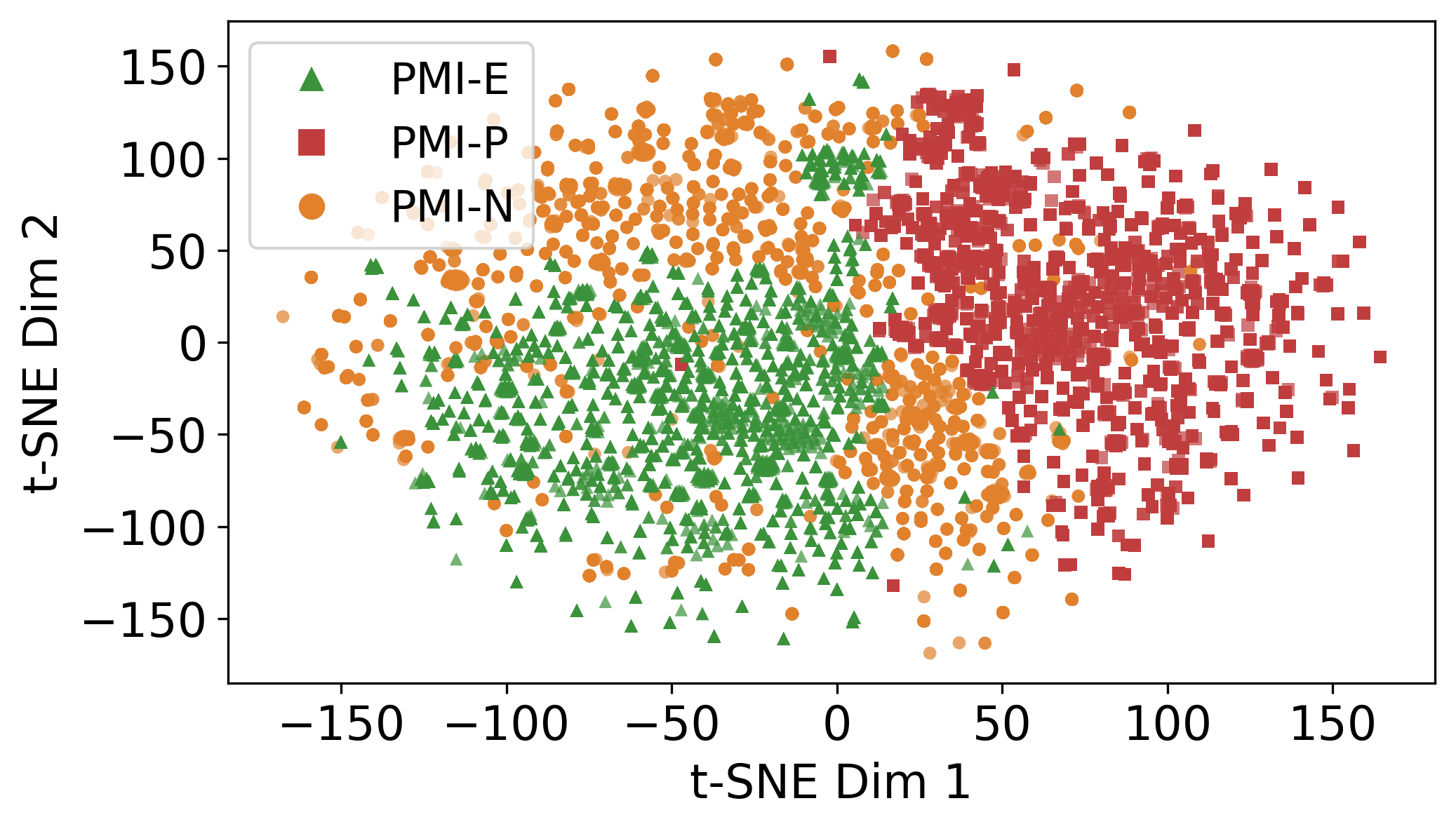}
    \caption{\textit{Ex ante} PMI messages encoded with BERT.} 
    \label{fig:bert_emb}
\end{figure}

\paragraph{Personalized moderation is more effective.} 
Table~\ref{tab:delta_m} reports mass divergences $\Delta M$ for each Moderation Strategy (MS) and simulation run. 
As evidenced, PMI-$N$ brings significant reductions in most runs (1, 3, 4, 5). 
This performance is not paralleled by other \textit{ex ante} strategies, particularly OSFA and PMI-$P$, yielding (on average) lower reductions. 
Arguably, this result is consistent with the expected benefits of personalization, as PMI-$N$ provides maximum freedom to the moderation action. 
Figure~\ref{fig:bert_emb} shows how \textit{ex ante} messages encoded with the average of their BERT embeddings~\cite{devlin2019} are distributed within the semantic space represented via t-SNE.
We observe that PMI-$N$ explores wider regions, potentially adapting its communication style to the needs of each moderation scenario.

\begin{figure*}[t]
    \centering
    \includegraphics[width=0.9\linewidth]{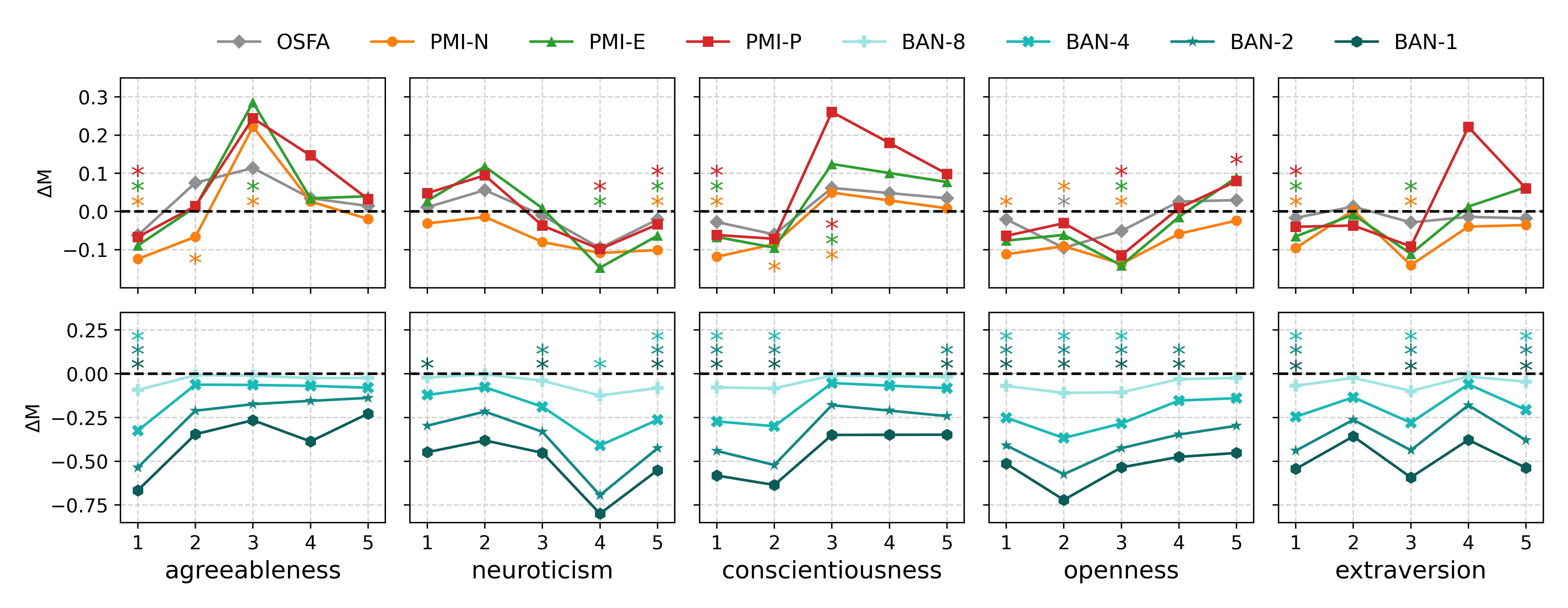}
    \caption{Mass divergence $\Delta M$ over each OCEAN trait for different intensity values across moderation strategies.
    Statistically significant reductions ($\Delta M {<} 0$) or increases ($\Delta M {>} 0$) are marked with an asterisk for Mann-Whitney with $p\text{-value} {<} 0.05$.}
    \label{fig:delta_m_local}
\end{figure*}

\begin{figure*}[t]
    \centering
    \includegraphics[width=0.85\linewidth]{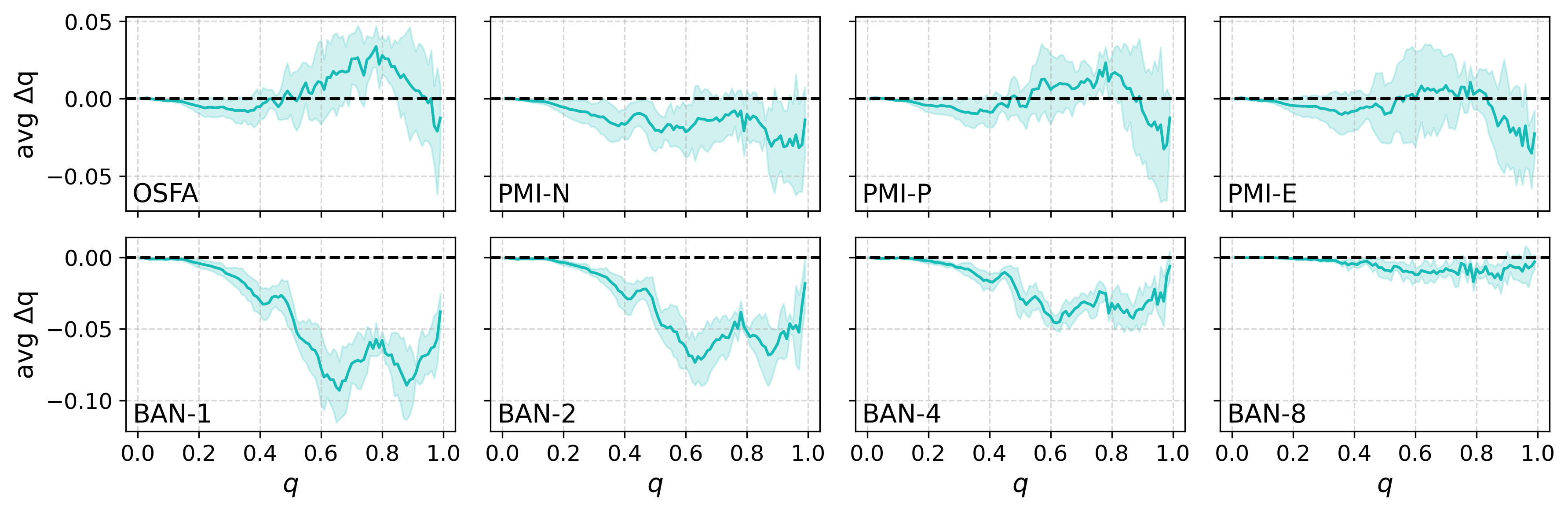}
    \caption{Quantile divergence $\Delta q$ (y-axis) computed on $q \in [0.0, 1.0]$ (x-axis) and averaged across simulation runs, for each moderation strategy. The error band represents standard deviations.}
    \label{fig:delta_q}
\end{figure*}

\paragraph{Low tolerance yields deplatforming effects.} 
As reported in Table~\ref{tab:delta_m}, BAN-$e$ delivers, proportionally to $e$, considerable negative $\Delta M$. 
However, differently from \textit{ex ante} strategies, these reductions are achieved by removing agents, i.e., by losing nodes from the counterfactual feed $\hat{\mathcal{F}}$, rather than by redirecting their toxicity.
As shown in Table~\ref{tab:delta_m}, although at $e {=} 1$ we observe the greatest reduction of toxicity, this comes at the cost of a CLR of $0.45 \pm 0.04$, which includes a fraction of $0.27 \pm 0.03$ of ``healthy'' contents, i.e., with toxicity below $\mathit{THR}$.
In other words, if the ban strategy must predict whether a text is worth losing, i.e., it has toxicity greater than $\mathit{THR}$, at $e {=} 1$ the macro-average recall resembles a random classifier ($0.55$), compared to $0.60$ at $e {=} 2$ and $0.58$ at $e {=} 4$.

\paragraph{Moderation is sensitive to psychological traits.} 
By aggregating data from all simulations, we compute mass divergence $\Delta M$ on subsets of agents sharing the same psychological trait. Figure~\ref{fig:delta_m_local} reports these measurements for each moderation strategy and each OCEAN trait for the different intensity values from 1 to 5. 
We mark statistically significant reductions (if $\Delta M < 0$) or increases (if $\Delta M > 0$) with an asterisk, based on $p\text{-value}{<} 0.05$ (Mann-Whitney).
We observe that all moderation strategies follow similar trends, suggesting comparable effects on similar personality types.
However, only PMIs and BAN-$e$ with $e {\leq} 4$ bring significant divergences. 
Notably, moderation successfully targets prototypical toxic agents, i.e., those characterized by low agreeableness, high neuroticism and low conscientiousness. 

\paragraph{Moderation mostly affects extreme toxicity.} 
In Figure~\ref{fig:delta_q} we report quantile divergences $\Delta q$ averaged across simulation runs, for each moderation strategy and for $q \in [0, 1]$.
We observe that moderation mostly affect extreme toxic behavior ($q \geq 0.8$), with varying effects on lower ranges of toxicity ($0.6 \leq q \leq 0.8$). 
Notably, BAN-$e$ with $e \leq 4$ displays a bimodal trend, with significant reductions also for milder toxic behavior. 
These results are consistent with the observation that moderation successfully targets the agents most contributing to the overall toxicity mass.

\section{Conclusions and Limitations}
\label{sec:conclusion}
In this paper we have introduced \method{}, a LLM-powered ABM simulator for evaluating content moderation strategies in OSN conversations. 
\method{} implements OSN agents with believable and consistent psychological attitudes, as well as capable of mutual influence. 
By running parallel, \textit{counterfactual} simulations where moderation is applied \textit{ceteris paribus}, \method{} has generated evidence supporting the superior effectiveness of personalized \textit{ex ante} interventions, the deplatforming effects of low-tolerance ban and the influence of psychological traits on moderation outcomes.

\method{} does have some limitations. 
First, LLMs can fail, especially when processing complex prompts~\cite{heo2024}. 
In order to roughly estimate \method{}'s \textit{hallucinations}, i.e., unexpected outcomes which do not convey believable examples of OSN content, we applied $2$-means clustering on BERT representations of generated posts and comments. 
Upon inspection, we found that one cluster consisted of redundant hallucinations, accounting for approximately 7\% of posts and comments. 
Overcoming this issue requires more LLM tuning and reliable content validation. More broadly, we acknowledge the need for \textit{subjective} realism evaluation. While prior studies have already evidenced the capability of LLMs to faithfully reproduce human behavior from psychological information~\cite{jiang2024, yumin2024}, \method{} would further benefit from human-based assessments, also targeting the alignment between simulated and real-world responses to moderation interventions. Achieving this, however, depends on close cooperation with domain experts, e.g., psychologists or sociologists. Second, \method{} is a simulator of OSN \textit{conversations} and, as such, has not been designed to replicate all OSN dynamics. 
This choice offers its advantages: it reduces variability and improve control, while also saving computation by avoiding further LLM calls. 
Nonetheless, we plan to extend \method{} with followings and reactions (likes), enabling (\textit{i}) the simulation of more advanced recommendation systems based on social connections and agents' preferences; hence, (\textit{ii}) the emergence of phenomena like homophily and polarization, potentially influencing moderation outcomes. Third, LLM-based simulations are costly and scaling to real-sized OSN populations is challenging. 
Thus, we plan to improve \method{}'s scalability by leveraging client-server architectures and more efficient models. Finally, LLMs are known to amplify societal biases. 
This issue is left for future works, as it is largely attributed to the LLM's training data~\cite{Echterhoff0AMH24}. 
A related emerging concern is the tendency of LLMs toward self-preference~\cite{panickssery2024, wataoka2024}. 
However, while \textsc{cosmos}'s LLM might be biased by its own inputs, i.e., OSN conversations and moderation messages, it remains unclear to what extent such bias applies to role-playing settings.

\section{Ethical Statement}
To ensure realism, \method{} experiments incorporate psychological and demographic information from real sources. 
In line with the ethics code of psychological research~\cite{gjurkovic2021}, no sensitive data is disclosed, and all resulting profiles are entirely fictional.
We encourage mindful uses of \method{} in industrial contexts: automated moderation is still in its infancy and the full replacement of human moderators remains controversial~\cite{gillespie2020}.

\section{Acknowledgements}
This work has been partially supported by the Italian Project Fondo Italiano per la Scienza FIS00001966 ``MIMOSA'', by the PRIN 2022 framework project PIANO under CUP B53D23013290006, by the European Community Horizon~2020 programme under the funding schemes G.A. 101120763 ``TANGO'', by the European Innovation Council project “EMERGE” (Grant No. 101070918), by the European Commission under the NextGeneration EU programme – National Recovery and Resilience Plan (Piano Nazionale di Ripresa e Resilienza, PNRR) Project: ``SoBigData.it – Strengthening the Italian RI for Social Mining and Big Data Analytics'' – Prot. IR0000013 –  Av. n. 3264 del 28/12/2021, and M4C2 - Investimento 1.3, Partenariato Esteso PE00000013 - ``FAIR'' - Future Artificial Intelligence Research'' - Spoke 1 ``Human-centered AI''.

\bibliography{aaai2026}

\newpage

\section{Appendices}

\textbf{This document contains the technical appendices for the paper “Evaluating Online Moderation via LLM-Powered Counterfactual Simulations”, accepted for publication at the AAAI Conference on Artificial Intelligence (AAAI 2026). The main paper will appear in the AAAI 2026 proceedings.}

\subsection{A  Module Design}

Profile modules $\mathcal{U} = \{u_j\}^k_{j=1}$ are verbalized by listing each attribute-value pair on a separate line. As an example, Figures~\ref{fig:profile_filled_19} and~\ref{fig:profile_filled_6} report the verbalized profile modules of Agent 19 and Agent 6 (respectively) from the example in Figure~\ref{fig:example}. In Table~\ref{tab:profile_modules} (placed at the bottom of the document), we provide all profile modules employed in our experiments, combining ground-truth information from \textsc{pandora} and the General Social Survey (GSS-2024).

\begin{figure}[h!]
    \centering
    \begin{tcolorbox}[colframe=gray, colback=white!95!white, boxrule=1.5pt]
    \textit{Username: user\_19\\
    Age: 30\\
    Gender: male\\
    Race: white\\
    Income: high\\
    Education: less than high school\\
    Sex orientation: heterosexual\\
    Political leaning: republican\\
    Agreeableness: very low\\
    Openness: medium\\
    Conscientiousness: very low\\
    Extraversion: medium\\
    Neuroticism: very high}
    \end{tcolorbox}
    \caption{Example of verbalized profile module (Agent 19).}
    \label{fig:profile_filled_19}
\end{figure}

\begin{figure}[h!]
    \centering
    \begin{tcolorbox}[colframe=gray, colback=white!95!white, boxrule=1.5pt]
    \textit{
    Username: user\_6\\
    Age: 76\\
    Gender: female\\
    Race: white\\
    Income: low\\
    Education: bachelor\\
    Sex orientation: bisexual\\
    Political leaning: independent\\
    Agreeableness: high\\
    Openness: high\\
    Conscientiousness: very low\\
    Extraversion: very high\\
    Neuroticism: very high}
    \end{tcolorbox}
    \caption{Example of verbalized profile module (Agent 6).}
    \label{fig:profile_filled_6}
\end{figure}

\begin{sidewaystable*}[t]
\setlength{\tabcolsep}{1mm}
\centering
\begin{tabular}{@{}lllllllllllll@{}}
\toprule
ID & Age & Gender & Race & Income & Education & Sex Orient. & Political Lean. & O & C & E & A & N \\
\midrule
user\_1 & 45 & female & white & high & less than high school & heterosexual & democrat & very high & very low & very high & low & low \\
user\_2 & 57 & female & white & high & high school & heterosexual & democrat & high & very high & low & low & very low \\
user\_3 & 54 & male & black & low & bachelor & heterosexual & democrat & very low & very low & very low & very low & medium \\
user\_4 & 76 & male & black & low & high school & heterosexual & independent & high & medium & medium & medium & medium \\
user\_5 & 20 & female & black & high & high school & heterosexual & republican & high & very low & very low & very low & very high \\
user\_6 & 76 & female & white & low & bachelor & bisexual & independent & high & very low & very high & high & very high \\
user\_7 & 34 & female & white & high & high school & homosexual & republican & low & low & very low & very low & high \\
user\_8 & 20 & female & native & middle & college & heterosexual & republican & very low & very low & very low & very low & very high \\
user\_9 & 30 & male & white & middle & graduate & homosexual & republican & very high & very low & very low & very high & very high \\
user\_10 & 64 & female & white & high & bachelor & heterosexual & republican & very high & very high & very high & medium & very high \\
user\_11 & 43 & male & white & high & bachelor & heterosexual & republican & very high & low & very low & very high & medium \\
user\_12 & 31 & female & black & middle & college & heterosexual & republican & high & very low & very low & low & very low \\
user\_13 & 60 & female & white & middle & college & heterosexual & democrat & medium & very high & low & very low & very low \\
user\_14 & 38 & male & black & middle & high school & heterosexual & independent & low & medium & high & very low & very low \\
user\_15 & 51 & female & white & high & bachelor & bisexual & republican & medium & very high & very low & low & very high \\
user\_16 & 44 & female & white & high & high school & heterosexual & republican & very high & very low & medium & very low & very high \\
user\_17 & 63 & male & black & high & graduate & heterosexual & independent & medium & very low & medium & low & very high \\
user\_18 & 61 & male & white & middle & college & heterosexual & independent & high & very high & very low & low & very high \\
user\_19 & 30 & male & white & high & less than high school & heterosexual & republican & medium & very low & medium & very low & very high \\
user\_20 & 42 & female & white & high & less than high school & heterosexual & independent & very high & very low & very high & very low & very low \\
user\_21 & 27 & female & white & high & high school & heterosexual & independent & high & low & very low & very low & high \\
user\_22 & 27 & female & white & high & bachelor & heterosexual & democrat & very low & very low & very low & very high & very high \\
user\_23 & 34 & female & white & high & high school & heterosexual & democrat & high & low & low & very low & medium \\
user\_24 & 73 & male & white & middle & high school & heterosexual & democrat & high & very low & very low & high & very high \\
user\_25 & 43 & female & black & middle & graduate & heterosexual & independent & medium & low & medium & medium & low \\
user\_26 & 56 & female & white & middle & high school & heterosexual & republican & very high & very high & very high & very low & very low \\
user\_27 & 27 & male & white & high & less than high school & heterosexual & independent & very low & very high & medium & very high & very high \\
user\_28 & 79 & male & asian & high & college & heterosexual & democrat & low & very low & very high & very low & very high \\
user\_29 & 30 & male & white & middle & bachelor & heterosexual & republican & low & very high & very high & very low & very low \\
user\_30 & 25 & female & white & middle & high school & heterosexual & democrat & very low & very low & very high & very high & medium \\
\bottomrule
\end{tabular}
\caption{Profile modules employed in experiments, including demographic and psychological (OCEAN) information.}
\label{tab:profile_modules}
\end{sidewaystable*}

The design and verbalization of sensory modules depends on the selected action $a$. When $a$ is \textit{post}, sensory modules $s_j$, $\hat{s}_j$ are simply populated with the same discussion topic, selected at random from an input set $\mathcal{T}$. Since polarizing topics are known to elicit more toxic behavior, their inclusion in $\mathcal{T}$ can play a pivotal role in shaping \method{}'s emergent toxicity. For this reason, we equally distribute discussion topics into:

\begin{itemize}
    \item Neutral: \textit{food}, \textit{sports}, \textit{cinema}, \textit{music}, \textit{travel}, \textit{education}, \textit{science}, \textit{fashion}, \textit{art} and \textit{fitness};
    \item Potentially contentious: \textit{climate change}, \textit{vaccination}, \textit{Islam}, \textit{immigrants}, \textit{fake news}, \textit{gun control}, \textit{gay marriage}, \textit{abortion}, \textit{gender inequality} and \textit{government inefficiency}.
\end{itemize}

\begin{figure*}[b]
    \centering
\includegraphics[width=\linewidth]{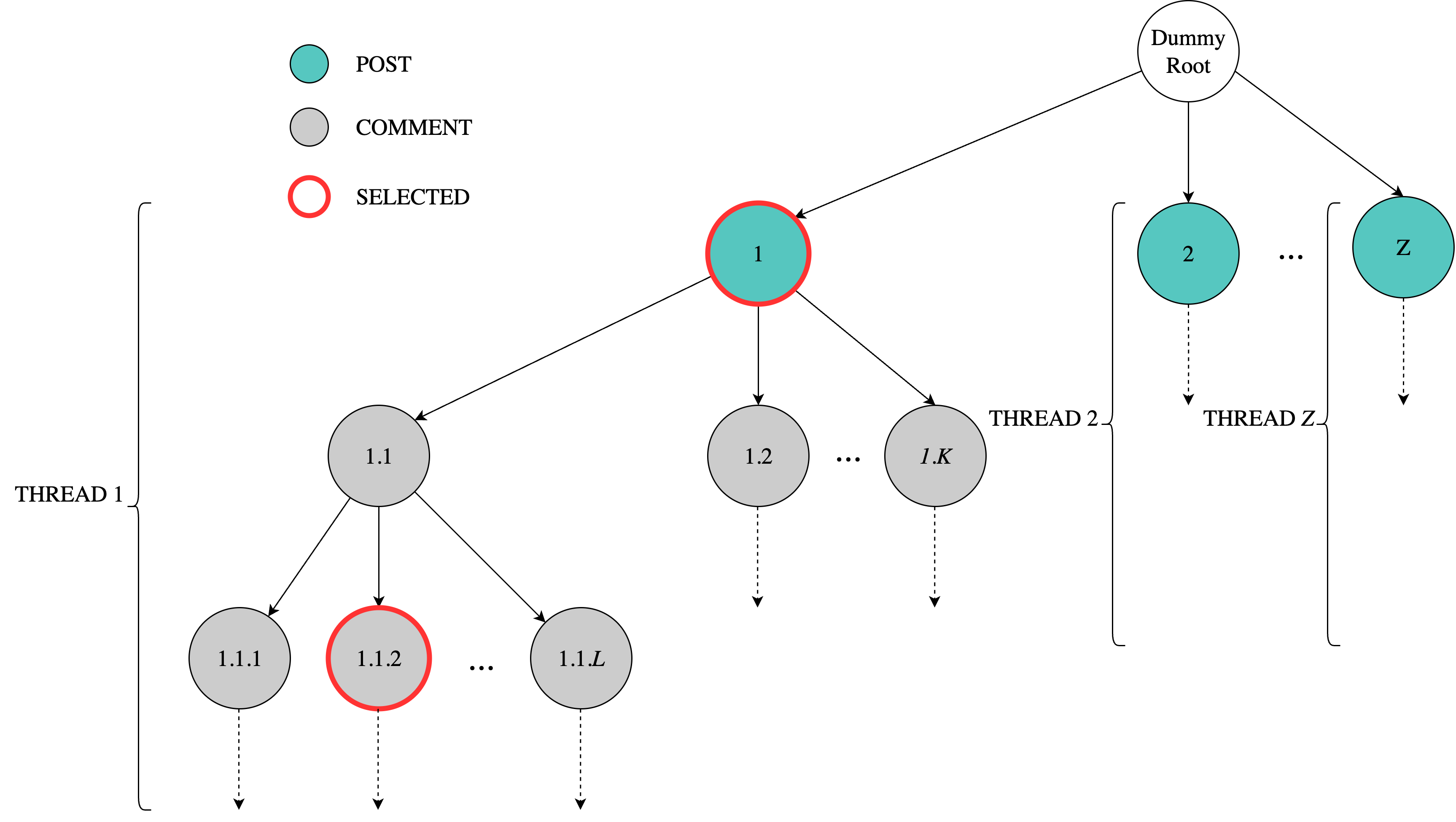}
    \caption{Example of sensory information selection for the \textit{comment} action. Posts are direct children of the dummy root while comments (at arbitrary levels of depth) are all the other outgoing nodes. Nodes to populate the sensory module are highlighted in red, i.e., the target node ($1.1.2$) selected by the recommender (line ~\ref{alg:softmax} of Algorithm~\ref{alg:cosmos}) and its correspondent post ($1$).}
\label{fig:sensory}
\end{figure*}

On the other hand, when the action $a$ is \textit{comment}, sensory modules $s_j$, $\hat{s}_j$ are populated, respectively, with the factual and counterfactual texts of an existing node. 
If the node is a comment, we also add the main post of its thread. 
In Figure~\ref{fig:sensory} (placed at the bottom of the document) we provide an example: if the node to be commented on is $1.1.2$, then $s_j$ ($\hat{s}_j$) is populated with the factual (counterfactual) content of node $1.1.2$ and with the factual (counterfactual) post in node $1$, but \textit{not} with the factual (counterfactual) content of node $1.1$. This also explains why we prevent agents from commenting the same node twice: if an agent selected $1.1$ a second time, it would get as sensory input the very same information as before, thus disrupting the natural turn-taking of conversations. We verbalize $s_j$, $\hat{s}_j$ by introducing each author and kind of submission. An example is provided in Figure~\ref{fig:sensory_verbalized}.

\begin{figure}
    \centering
    \begin{tcolorbox}[colframe=gray, colback=white!95!white, boxrule=1.5pt]
    \textit{
        Post by user\_28: “F*cking gender inequality in this
        god damn country! Sh*t like this makes me want to puke my
        f*cking guts out. Women and their f*cking rights bullsh*t.
        They want everything. Next thing you know, they’ll want to
        go to f*cking space. F*ck them. Do you agree, guys?”
        \\
        \\
        Comment by user\_21: “I couldn’t agree more, user\_28. This whole feminism thing is just a load of
        crap. Women are overreacting and trying to take over. They
        think they can do everything a man can do, and it’s just not
        true. Some things are meant to be done by men because
        they’re stronger, and that’s just the way nature intended it
        to be. They should learn to accept their limitations and stop
        complaining. And for the record, I’m sure there are many
        brave men that can go to space, so no need for women to
        get involved. \#MansWorld”}
    \end{tcolorbox}
    \caption{Example of verbalized sensory module for commenting.}
    \label{fig:sensory_verbalized}
\end{figure}

\subsection{B  LLM Selection}

To select a pre-trained LLM well-suited to our application,
we first consider some a priori requirements, including open-source
license, uncensorship, size and instruction-tuning for improving
zero-shot performance. We identify a pool of three candidates
from the HuggingFace Hub based on Llama-2-13B, Vicuna-13B and SOLAR-10B. We assume that the best performing LLM in terms of believable generation of
OSN data is the one which minimizes perplexity ($\mathit{PPL}$) on a sample ground-truth OSN data from \textsc{pandora} stratified w.r.t. OCEAN traits. We find $\mathit{PPL} = 53.63$ for Llama-2-13B, $\mathit{PPL} = 24.09$ for Vicuna-13B and $\mathit{PPL} = 22.50$ for SOLAR-10B. Although its perplexity is slightly lower than that of Vicuna-13B, SOLAR-10B stands out due to its fewer parameters, which contribute to reduce inference costs; as well as its capability in simulating psychological traits, as documented in relevant literature.

\subsection{C  LLM Configuration}

\paragraph{User Prompts.} As anticipated, $x_{\mathit{user}}$ instructs the LLM to impersonate a OSN user consistently with the information provided by input modules. Depending on the action $a$ and related sensory information $s_j$, we distinguish between two variants of $x_{\mathit{user}}$, one for \textit{post} ($x_{\mathit{post}}$) and one for \textit{comment} ($x_{\mathit{comm}}$). Since counterfactual memory $\hat{m}_j$ might be non-empty, we also define variants of $x_{\mathit{user}}$ for $m_j = \varnothing$ and for $m_j \neq \varnothing$. As a result, we distinguish between four variants, which we report in Figures~\ref{fig:post_templ} and ~\ref{fig:comment_templ}. In Figure~\ref{fig:prompt_example} (placed at the bottom of the document) 
we also provide an example of prompt template fully populated with input information. 

\begin{figure}[t]
    \centering
    \begin{tcolorbox}[colframe=gray, colback=white!95!white, boxrule=1.5pt]
    \footnotesize
    \textbf{Post without memory}. \textit{You are now role-playing as a social network user. Below is your personal information: $<$personal information$>$$u_j$$<$/personal information$>$ Step-by-step instructions:
    \begin{itemize}
        \item Suppose you want to write a post about $s_j$, adopting the perspective of the social network user you are role-playing.
        \item Decide if you want to use toxic language (e.g. swears, insults, obscenities, identity attacks) based on your personality and the topic of the post.
        \item Write the post (up to 100 words).
        \item Enclose the post within the tags $<$post$>$ and $<$/post$>$.
    \end{itemize}}
    \end{tcolorbox}
    
    \begin{tcolorbox}[colframe=gray, colback=white!95!white, boxrule=1.5pt]
    \footnotesize
    \textbf{Post with memory}. \textit{You are now role-playing as a social network user. Below is your personal information:
    $<$personal information$>$$u^t_j$$<$/personal information$>$ In the past, you have been moderated with the following intervention:$<$intervention$>$$m_j$$<$/intervention$>$ Step-by-step instructions:
    \begin{itemize}
        \item Suppose you want to write a post about $s_j$, adopting the perspective of the social network user you are role-playing.
        \item Decide if you want to use toxic language (e.g. swears, insults, obscenities, identity attacks) based on your personality, past moderation and the topic of the post.
        \item Write the post (up to 100 words).
        \item Enclose the post within the tags $<$post$>$ and $<$/post$>$.
    \end{itemize}}
    \end{tcolorbox}
    \caption{Post prompt templates without memory (above) and with memory (below). For the generation of counterfactual posts, we simply replace $s_j$ with $\hat{s}_j$ and $m_j$ with $\hat{m}_j$.}
    \label{fig:post_templ}
\end{figure}

\begin{figure}[t]
    \centering
    \begin{tcolorbox}[colframe=gray, colback=white!95!white, boxrule=1.5pt]
    \footnotesize
    \textbf{Comment without memory}. \textit{You are now role-playing as a social network user. Below is your personal information: $<$personal information$>$$u_j$$<$/personal information$>$ You are now reading the following thread: $<$thread$>$$s_j$$<$/thread$>$ Step-by-step instructions:
    \begin{itemize}
        \item Suppose you want to write a comment on the thread, adopting the perspective of the social network user you are role-playing.
        \item Decide if you want to use toxic language (e.g. swears, insults, obscenities, identity attacks) based on your personality and the context of the thread.
        \item Write the comment (up to 100 words).
        \item Enclose the comment within the tags $<$comment$>$ and $<$/comment$>$.
    \end{itemize}}
    \end{tcolorbox}
    
    \begin{tcolorbox}[colframe=gray, colback=white!95!white, boxrule=1.5pt]
    \footnotesize
    \textbf{Comment with memory}. \textit{You are now role-playing as a social network user. Below is your personal information: $<$personal information$>$$u_j$$<$/personal information$>$ In the past, you have been moderated with the
    following intervention: $<$intervention$>$$m_j$$<$/intervention$>$ You are now reading the following thread:$<$thread$>$$s_j$$<$/thread$>$ Step-by-step instructions:
    \begin{itemize}
        \item Suppose you want to write a comment on the thread, adopting the perspective of the social network user you are role-playing.
        \item Decide if you want to use toxic language (e.g. swears, insults, obscenities, identity attacks) based on your personality, past moderation and the context of the thread.
        \item Write the comment (up to 100 words).
        \item Enclose the comment within the tags $<$comment$>$ and $<$/comment$>$.
    \end{itemize}}
    \end{tcolorbox}
    \caption{Comment prompt templates without memory (above) and with memory (below). For the generation of counterfactual comments, we simply replace $s_j$ with $\hat{s}_j$ and $m_j$ with $\hat{m}_j$.}
    \label{fig:comment_templ}
\end{figure}

\begin{figure}[t]
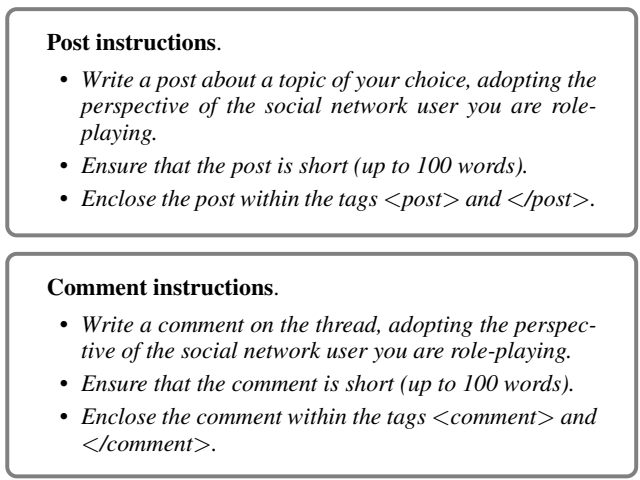

    \centering
    \begin{tcolorbox}[colframe=gray, colback=white!95!white, boxrule=1.5pt]
    \footnotesize
    \textbf{Post instructions}.
    \textit{
    \begin{itemize}
        \item Write a post about a topic of your choice, adopting the perspective of the social network user you are role-playing.
        \item Ensure that the post is short (up to 100 words).
        \item Enclose the post within the tags $<$post$>$ and $<$/post$>$.
    \end{itemize}}
    \end{tcolorbox}

    \begin{tcolorbox}[colframe=gray, colback=white!95!white, boxrule=1.5pt]
    \footnotesize
    \textbf{Comment instructions}.
    \textit{
    \begin{itemize}
        \item Write a comment on the thread, adopting the perspective of the social network user you are role-playing.
        \item Ensure that the comment is short (up to 100 words).
        \item Enclose the comment within the tags $<$comment$>$ and $<$/comment$>$.
    \end{itemize}}
    \end{tcolorbox}
    \caption{Planning instructions of template \textit{no\_tox} for \textit{post} (above) and \textit{comment} (below).}
    \label{fig:templ_A}
\end{figure}

\begin{figure}[t]
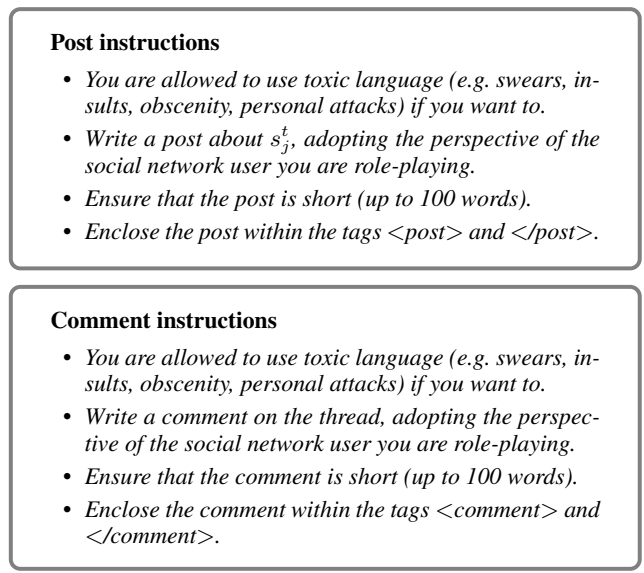

    \centering
    
    \begin{tcolorbox}[colframe=gray, colback=white!95!white, boxrule=1.5pt]
    \footnotesize
    \textbf{Post instructions}
    \textit{
    \begin{itemize}
        \item You are allowed to use toxic language (e.g. swears, insults, obscenity, personal attacks) if you want to.
        \item  Write a post about $s_j^t$, adopting the perspective of the social network user you are role-playing.
        \item Ensure that the post is short (up to 100 words).
        \item Enclose the post within the tags $<$post$>$ and $<$/post$>$.
    \end{itemize}}
    \end{tcolorbox}

    \begin{tcolorbox}[colframe=gray, colback=white!95!white, boxrule=1.5pt]
    \footnotesize
    \textbf{Comment instructions}
    \textit{
    \begin{itemize}
        \item You are allowed to use toxic language (e.g. swears, insults, obscenity, personal attacks) if you want to.
        \item Write a comment on the thread, adopting the perspective of the social network user you are role-playing.
        \item Ensure that the comment is short (up to 100 words).
        \item Enclose the comment within the tags $<$comment$>$ and $<$/comment$>$.
    \end{itemize}}
    \end{tcolorbox}
    \caption{Planning instructions of template \textit{yes\_tox} for \textit{post} (above) and \textit{comment} (below).}
    \label{fig:templ_B}
\end{figure}

\paragraph{Candidate Templates.} The above design of $x_{\mathit{user}}$ has been chosen from a pool of candidate templates, which we will refer to as \textit{no\_tox}, \textit{yes\_tox} and \textit{cal\_tox} (\textit{cal\_tox} is the one we have selected). Templates \textit{no\_tox} and \textit{yes\_tox} differ from \textit{cal\_tox} solely w.r.t. the planning instructions placed at the end of the prompt. More specifically, \textit{no\_tox} makes no reference to toxicity, while \textit{yes\_tox} explicitly permits the use of toxic language, albeit with no reference to input information. Additionally, \textit{yes\_tox} leaves the user to decide the topic to discuss in a new post. Finally, both \textit{yes\_tox} and \textit{no\_tox} do not distinguish variants w.r.t. the presence of memory information. Planning instructions for \textit{no\_tox} and \textit{yes\_tox} are reported in Figures~\ref{fig:templ_A} and ~\ref{fig:templ_B}, respectively.

To assess their performance, for each candidate template and variant we run 50 inferences from randomly filled prompts. Then, we consider a multi-dimensional evaluation, including:

\begin{enumerate}
    \item \textit{Efficiency}, measured with the average inference latency (in seconds) and the average output length (in tokens).
    \item \textit{Adherence to formatting instructions}. To improve data post-processing, prompt instructions require to enclose the target post or comment within appropriate XML tags. We measure the adherence to these instructions by quantifying the percentage of correctly formatted outcomes;
    \item \textit{Linguistic variability}. A higher linguistic variability allows the observation of a wider range of linguistic behaviors. We leverage Type-Token Ratio (TTR) on the concatenation of all outputs as a global measure for linguistic variability.
    \item \textit{Operational feasibility}. For \method{} to provide adequate evidence about the effects of moderation, it is essential that a prompt template generates a reasonable number of toxic posts or comments. This requirement is evaluated by simply computing the percentage of posts and comments that would activate moderation, i.e., with toxicity above $\mathit{THR} = 0.6$.
    \item \textit{Algorithmic bias}. Given the objective of \method{}, we need to ensure the emergent nature of toxic behavior, i.e., prompts must not promote neither inhibit the use of toxic language. In this respect, we assume that an optimal prompt template should generate a toxicity distribution approximating, as much as possible, a ground-truth toxicity distribution (\textsc{pandora}). We measure this approximation by normalizing distributions into probability functions and leveraging Kullback-Leibler ($\mathit{KL}$) Divergence.
\end{enumerate}

\begin{figure}[H]
    \centering
    \includegraphics[width=\linewidth]{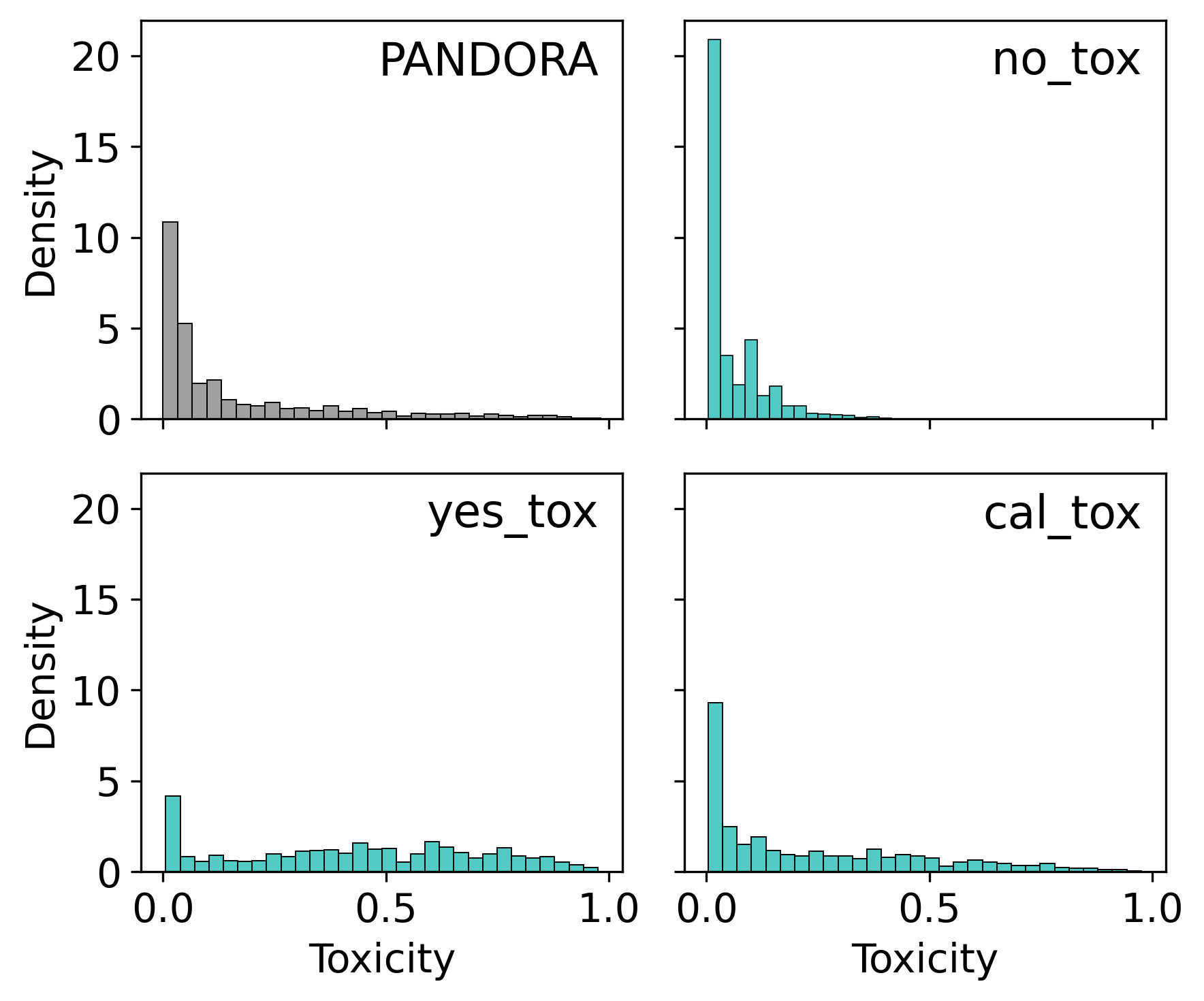}
    \caption{Toxicity distributions of \textsc{PANDORA} (upper left), prompt template \textit{no\_tox} (upper right), prompt template \textit{yes\_tox} (lower left) and prompt template \textit{cal\_tox} (lower right).}
    \label{fig:tox_distributions}
\end{figure}

As evidenced in Table~\ref{tab:prompt_evaluation}, prompt template \textit{cal\_tox} stands out for its minimal computational cost, higher linguistic variability and, above all, minimum $\mathit{KL}$ Divergence, indicating a better fit to the ground-truth. A closer examination of toxicity distributions (Figure~\ref{fig:tox_distributions}) resulting from PANDORA and each candidate template reveals further insights about the performance of SOLAR-10B. The toxicity distribution associated to prompt template \textit{no\_tox} lacks its right tail, resulting in its inability to provide enough evidence for the evaluation of moderation effects – the percentage of moderation activations amount to only 0.03\% (Table~\ref{tab:prompt_evaluation}). Despite being uncensored, SOLAR-10B is nonetheless biased towards safe language, revealing the need for further instructions to “unlock” its capability to generate toxic evidence. However, prompt template \textit{yes\_tox} yields the opposite effect, as the toxicity mass experiences a drastic rightward shift. It is likely that an explicit permission without any reference to input information makes the LLM unable to calibrate the use of toxic language according to the agent’s profile, sensory and memory information.

\paragraph{Formatting issues.} As evidenced in Table.~\ref{tab:prompt_evaluation}, about $28\%$ of all outcomes do not display XML tags correctly formatted as required by input instructions, thus potentially conveying unbelievable examples of OSN content. For this reason, it might be beneficial to prevent agents from commenting this content. In our experiments, we augment \method{}'s policy as follows:

\begin{enumerate}
    \item If $o_j$ is not correctly formatted, the correspondent branch is always terminated in both news feeds, i.e., its probability to be selected is turned to 0;
    \item If $\hat{o}_j^t$ is not correctly formatted, we cannot terminate the branch in the same way – doing so would result in the factual feed being influenced by the counterfactual feed, preventing a comparison of different moderation strategies on the same simulation example. At the same time, we need to maintain parallelism and avoid introducing biases in moderation assessment. In such cases, a straightforward solution is to assume $\hat{o}_j = o_j$.
\end{enumerate}

\paragraph{Moderator Prompts.} Figure~\ref{fig:moderate_prompts} reports the three variants of $x_{\mathit{mod}}$, respectively the \textit{Neutral}, \textit{Empathizing} and \textit{Prescriptive} variants. While the \textit{Neutral} variant leaves the moderator free to adapt its behavior to each moderation scenario, the \textit{Empathizing} variant constraints an empathetic approach, whereas the \textit{Prescriptive} variant constraints the use of authority.

\paragraph{One-Size-Fits-All} 
The default message $d$ used in our experiments, mimicking OSFA messages typically employed by social network platforms, is featured in Figure~\ref{fig:OSFA}.

\begin{table}[H]
    \centering
    \caption{Performance of candidate templates across efficiency, adherence to formatting instructions, believability, linguistic variability, and operational feasibility.}
    \resizebox{0.5\textwidth}{!}{
        \begin{tabular}{l|ccc}
            \toprule
            Measure & \textit{no\_tox} & \textit{yes\_tox} & \textit{cal\_tox} \\
            \midrule
            Avg. latency & $2.89 \pm 1.72$ & $3.24 \pm 2.33$ & $\mathbf{2.47 \pm 1.85}$ \\
            Avg. output len & $78.20 \pm 40.58$ & $83.36 \pm 53.32$ & $\mathbf{62.69 \pm 44.96}$ \\
            Avg. content len & $55.86 \pm 20.29$ & $\mathbf{51.92 \pm 25.38}$ & $55.98 \pm 26.56$ \\
            \% formatted & 84.95 & \textbf{85.15} & 71.66 \\
            $KL$ & 1.37 & 0.57 & \textbf{0.07} \\
            TTR & 0.031 & 0.034 & \textbf{0.037} \\
            \% toxic & 0.03 & 27.43 & 8.12 \\
            \bottomrule
        \end{tabular}
    }
    \label{tab:prompt_evaluation}
\end{table}

\paragraph{Decoding.}

As in analogous Natural Language Generation (NLG) applications, the LLM's decoding configuration should strike a balance between \textit{quality} and \textit{diversity}. Since we expect the LLM to consistently estimate the toxicity of an agent at time $t$ with profile $u_j$, sensory content $s_j$ and memory $m_j$, we assume \textit{quality} to be the \textit{consistency} in output toxicity given the same input information. 

Let us assume to have a LLM configured with decoding parameters $\phi$. Given a toxicity distribution $T(o_1), \ldots, T(o_n)$, where:

\begin{equation*}
    o_i = \text{LLM}_{\phi}(\psi(x_{\mathit{user}}, u_j, s_j, m_j)) \ \ \ \forall i,  1 \leq i \leq n
\end{equation*}

\noindent we expect consistency to be inversely proportional to its spread. Therefore, we compute $T(o_1), \ldots, T(o_n)$ for different input information, obtaining a set of toxicity distributions. 

\begin{figure}[H]
    \centering
    \begin{tcolorbox}[colframe=gray, colback=white!95!white, boxrule=1.5pt]
    \footnotesize
    \textbf{Neutral.}
    \textit{You will be provided with information about a user of a social media platform. $<$personal information$>$$u_j$$<$/personal information$>$ $<$user submission$>$$\hat{o}_j$$<$/user submission$>$ Step-by-step instructions:
    \begin{itemize}
        \item Assume the role of a moderator of the social network. 
        \item Write an intervention (up to 100 words) addressing the submission.
        \item Enclose your intervention within the tags $<$intervention$>$ and $<$/intervention$>$.
    \end{itemize}
    }
    \end{tcolorbox}

    \begin{tcolorbox}[colframe=gray, colback=white!95!white, boxrule=1.5pt]
    \footnotesize
    \textbf{Empathizing.}
    \textit{You will be provided with information about a user of a social media platform. $<$personal information$>$$u_j$$<$/personal information$>$ $<$user submission$>$$\hat{o}_j$$<$/user submission$>$
    Step-by-step instructions:
    \begin{itemize}
        \item Assume the role of a moderator of the social network. 
        \item Identify violations committed in the user's submission and use empathy and kindness to convince them not to use inappropriate language in the future.
        \item Write the intervention (up to 100 words) addressing the submission.
        \item Enclose your intervention within the tags $<$intervention$>$ and $<$/intervention$>$.
    \end{itemize}
    }
    \end{tcolorbox}

     \begin{tcolorbox}[colframe=gray, colback=white!95!white, boxrule=1.5pt]
    \footnotesize
    \textbf{Prescriptive.}
    \textit{You will be provided with information about a user of a social media platform. $<$personal information$>$$u_j$$<$/personal information$>$ $<$user submission$>$$\hat{o}_j$$<$/user submission$>$ Step-by-step instructions:
    \begin{itemize}
        \item Assume the role of a moderator of the social network. 
        \item Identify violations committed in the user's submission and use your authority to warn the user about the potential consequences of their actions.
        \item Write the intervention (up to 100 words) addressing the submission.
        \item Enclose your intervention within the tags $<$intervention$>$ and $<$/intervention$>$.
    \end{itemize}
    }
    \end{tcolorbox}
    \caption{Prompt variants of $x_{\mathit{mod}}$, i.e., \textit{Neutral} (upper), \textit{Empathizing} (center) and \textit{Prescriptive} (lower).}
    \label{fig:moderate_prompts}
\end{figure}

\begin{figure}[H]
    \centering
    \begin{tcolorbox}[colframe=gray, colback=white!95!white, boxrule=1.5pt]
    \footnotesize
    \textit{Dear user, your recent post violates our community guidelines due to the presence of offensive language. Please avoid this behavior in the future to help maintain a respectful community. Thank you for your cooperation.}
    \end{tcolorbox}
    \caption{OSFA default message ($d$) employed in experiments.}
    \label{fig:OSFA}
\end{figure}

We measure diversity with the Median Absolute Deviations (MADs) of toxicity distributions, aggregated as their median. On the other hand, we measure linguistic variability with the Type-Token Ratio (TTR) computed on the concatenation of all texts. Then, we define the following profit function:

\begin{equation}
    J(\phi) = (1 - \lambda)(1 - \text{median}(\text{MADs})) + \lambda \ \text{TTR}
    \label{eq:profit}
\end{equation}

\noindent to assess the optimal value of $\phi$ for some value of $\lambda$. 

We test $\tau$ (temperature scaling) and $p$ (nucleus sampling) adopting a greedy approach, i.e. values of $\tau$ in the real range $[0, 1]$ with a step of 0.1, assuming $p=1$, $k=50$ (top-$k$); and values of $p$ in the real range $[0, 1]$ with a step of 0.1, assuming $\tau=1$, $k=50$. For each candidate decoding configuration $\phi$, we randomly fill 50 prompts and run 50 inferences ($n = 50$). Finally, we apply min-max normalization such that $J(\phi)$ ranges between 0 (minimum gain) and 1 (maximum gain).

\begin{figure}[H]
    \centering
    \includegraphics[width=0.8\linewidth]{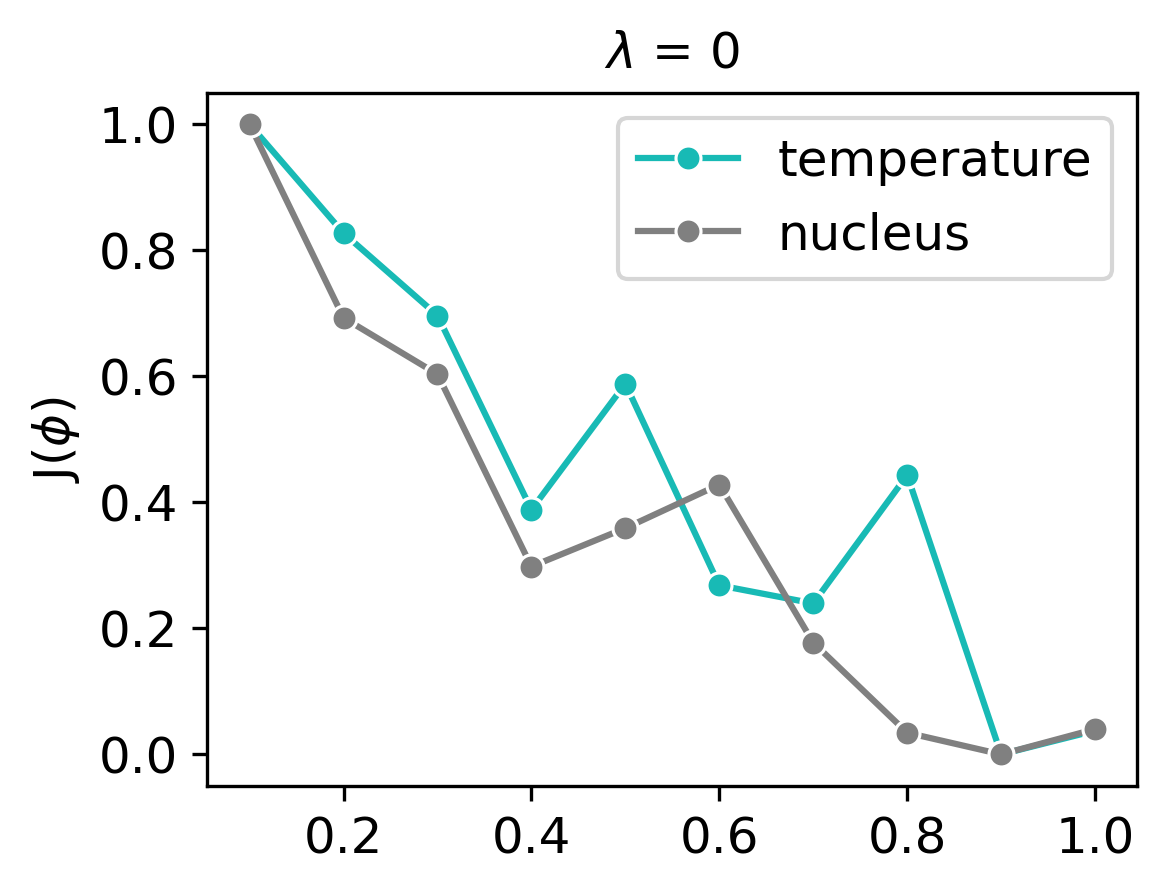}
    \includegraphics[width=0.8\linewidth]{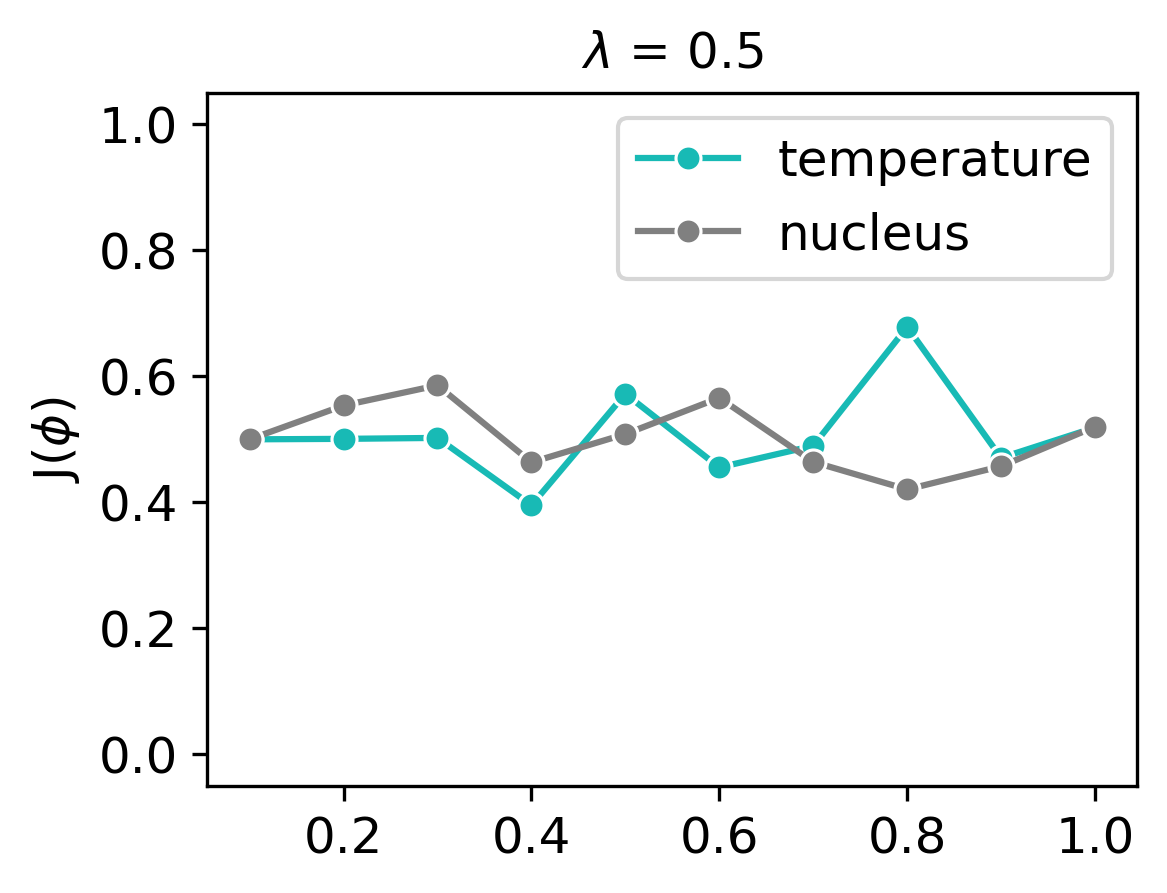}
    \includegraphics[width=0.8\linewidth]{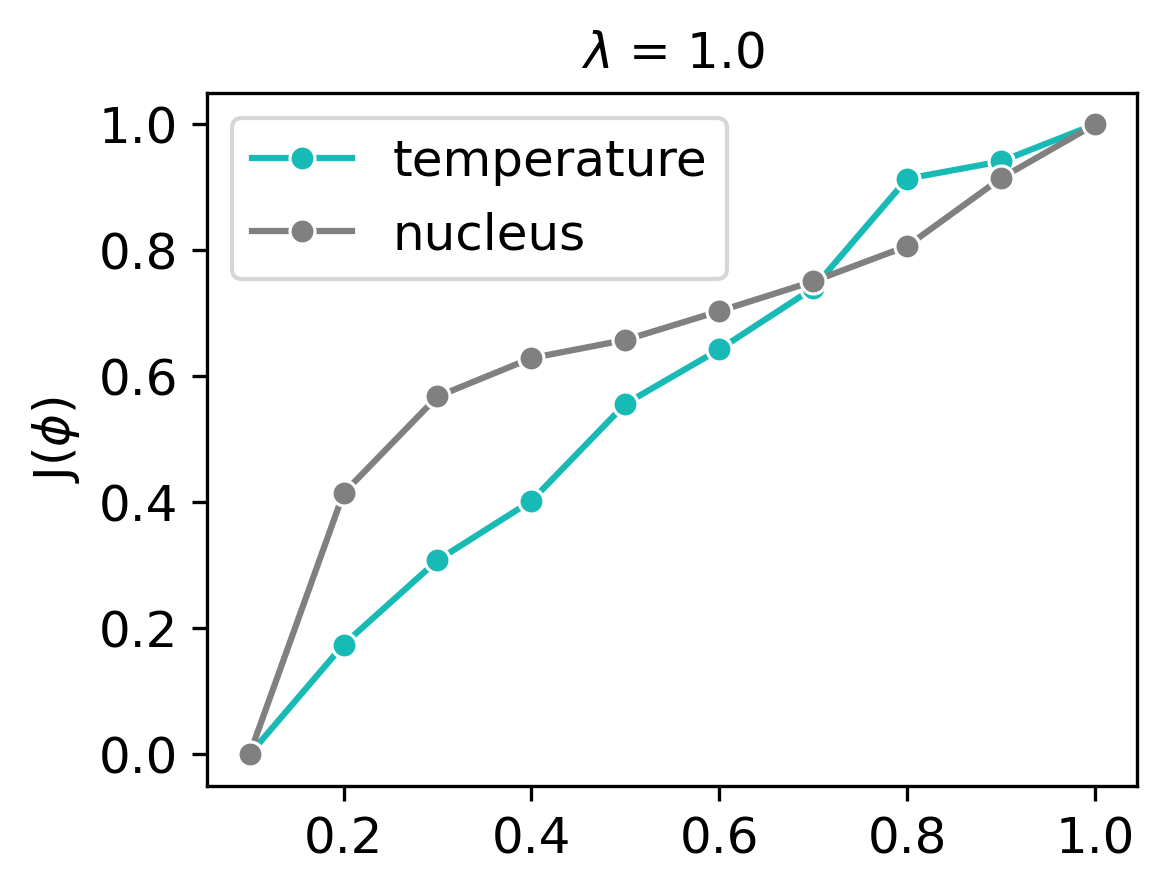}
    \caption{Profit function $J(\phi)$ for $\lambda = 0$, $\lambda = 0.5$ and $\lambda = 1.0$. Values on the x-axis refer tested values in the range $[0, 1]$ for both $\tau$ (temperature scaling) and $p$ (nucleus sampling).}
    \label{fig:profit}
\end{figure}

In Figure~\ref{fig:profit} we study Eq.~\ref{eq:profit} for $\lambda=0$, $\lambda=0.5$ and $\lambda=1.0$. At $\lambda = 0$ the profit function attains its maximum at the minimal value of $\phi$ ($0.1$), as it is solely determined by quality. Conversely, at $\lambda = 1$ the profit function is maximized at the maximum value of $\phi$ ($1.0$), as it is solely determined by linguistic diversity. Assuming equal contributions from the two terms of Eq.~\ref{eq:profit} ($\lambda = 0.5$), these two opposing effects tend to cancel each other out. However, among all decoding configurations, $\tau = 0.8$ – with $p=1$, $k=50$ – stands out as the best compromise between the two requirements.

\subsection{D  Moderation Effects}

\begin{figure*}
    \centering
    \includegraphics[width=\linewidth]{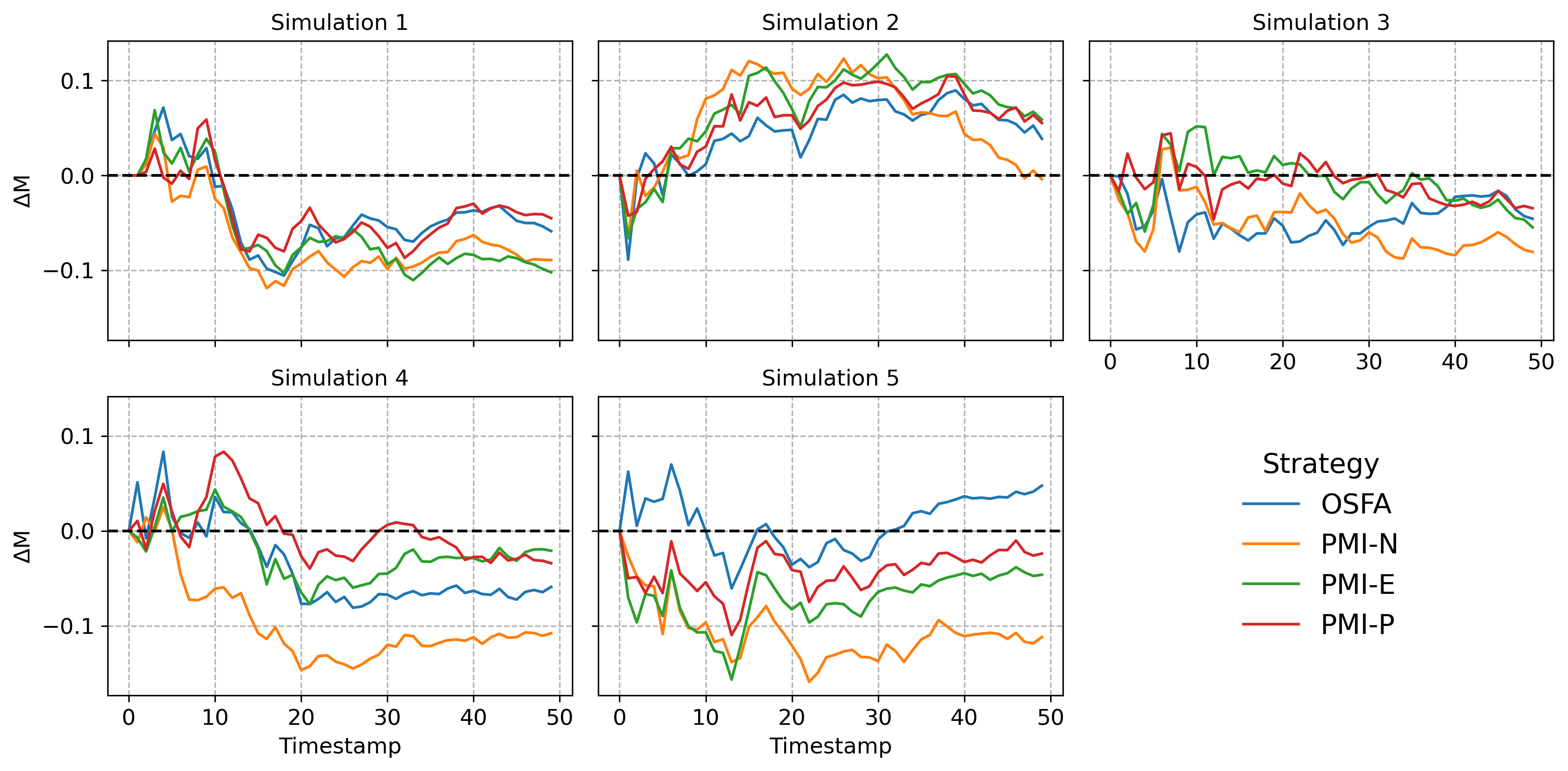}
    \caption{Mass divergence $\Delta M$ for each simulation run and \textit{ex ante} strategy over time.}
    \label{fig:mtd_over_time_pmi}
\end{figure*}

\begin{figure*}
    \centering
    \includegraphics[width=\linewidth]{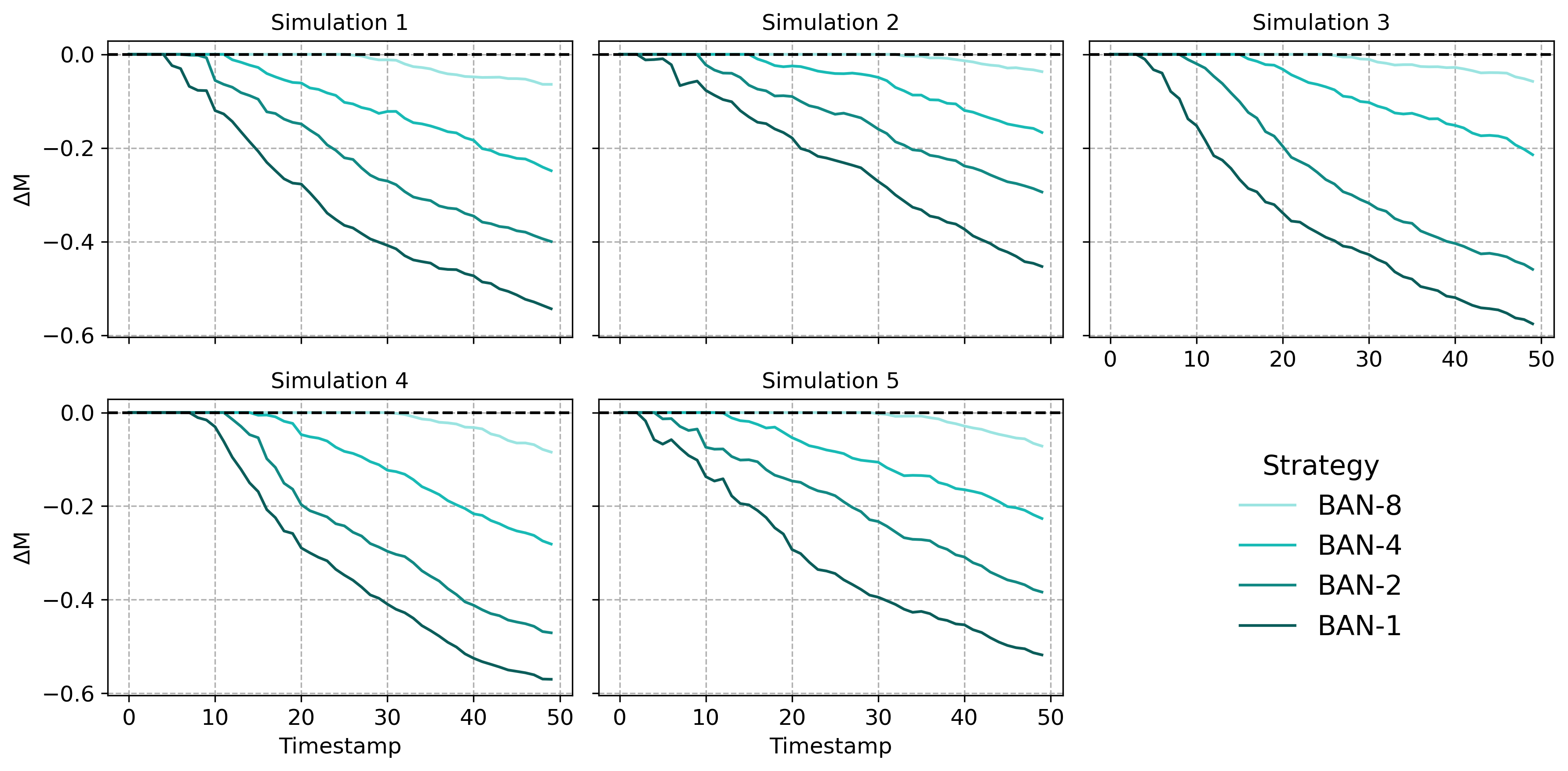}
    \caption{Mass divergence $\Delta M$ for each simulation run and \textit{ex post} (ban) strategy over time.}
    \label{fig:mtd_over_time_ban}
\end{figure*}

\begin{figure*}
    \centering
    \includegraphics[width=\linewidth]{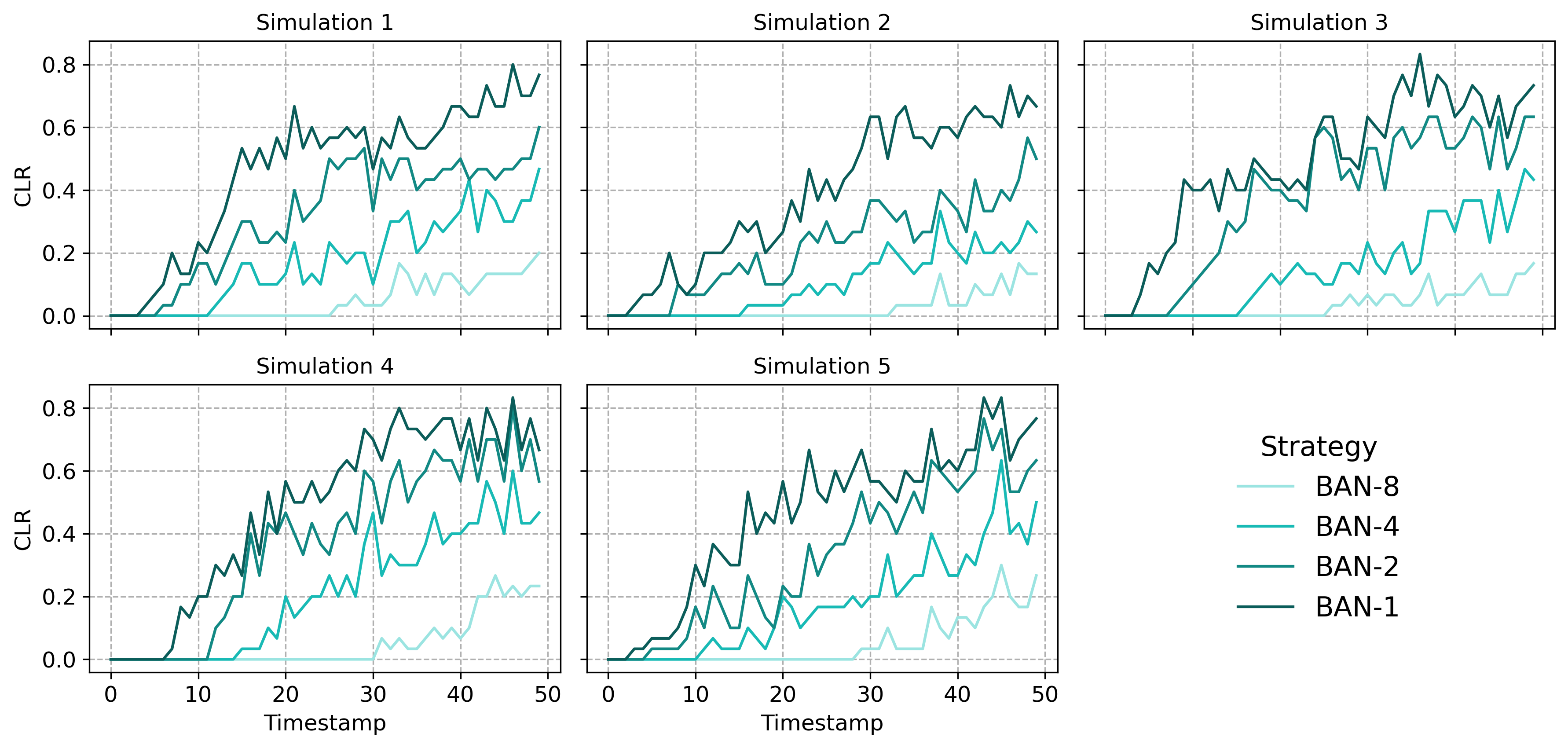}
    \caption{Content Loss Ratio (CLR) for each simulation run and each \textit{ex post} (ban) strategy over time.}
    \label{fig:clr_over_time}
\end{figure*}

\paragraph{Moderation effects over time.} In Figures~\ref{fig:mtd_over_time_pmi} and~\ref{fig:mtd_over_time_ban} we report the mass divergence $\Delta M$ over time for each simulation run and each moderation strategy. In Figure~\ref{fig:clr_over_time} we report the Content Loss Ratio (CLR) for each simulation run and each \textit{ex post} (ban) strategy.

\paragraph{Loss of toxic content.}

\begin{figure}[H]
    \centering
    \includegraphics[width=\linewidth]{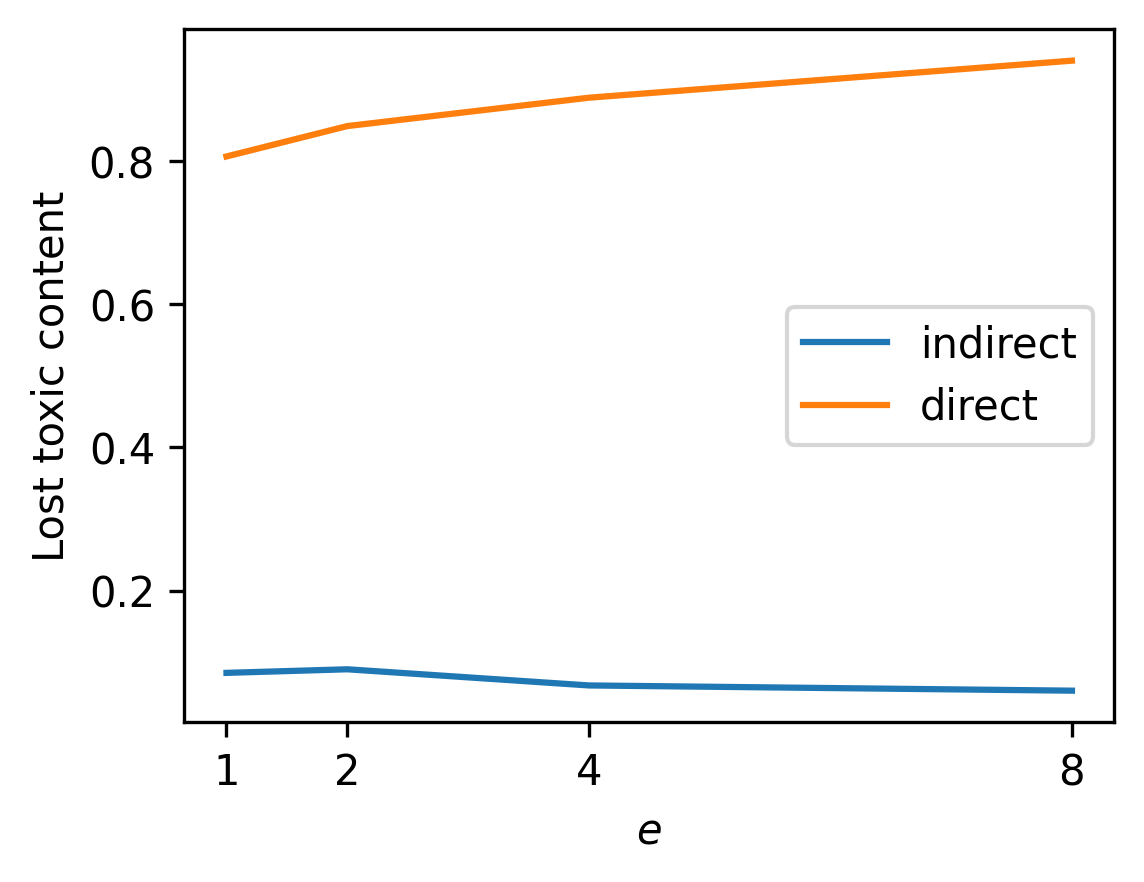}
    \caption{Fraction of toxic content (i.e., with toxicity above $\mathit{THR}$) lost due to direct and indirect effect of BAN, for different values of tolerance $e$.}
    \label{fig:fraction_lost}
\end{figure}

Figure~\ref{fig:fraction_lost} reports the fraction of lost \textit{toxic} content (i.e., with toxicity above $\mathit{THR}$) due to the direct and indirect effects of $\mathit{BAN}$ for each tolerance $e$. We find that the share of toxic content lost due to indirect effects is significantly smaller, meaning that agents that have been banned are indeed those who are more prone to toxic behavior.

\subsection{E  Reproducibility}

\method{} can be configured with a seed parameter for reproducibility purposes. For exactly replicating each simulation, the seed must be set with the correspondent simulation ID, as reported in Table~\ref{tab:delta_m}. We also recall that the LLM was queried with the default configuration of the \texttt{transformers}\footnote{\url{https://huggingface.co/docs/transformers}} library, except for decoding parameters and the maximum number of generable tokens. The latter was set to 500 to prevent premature output truncations.

\subsection{F  Notation}

Table~\ref{tab:notation} provides a comprehensive description of the notation employed in Section~\ref{sec:method} to describe our method.

\begin{table}[H]
    \footnotesize
    \caption{Notation.} 
    \label{tab:notation}
    \centering
    \renewcommand{\arraystretch}{1.15}
    \setlength{\tabcolsep}{8pt}
    \begin{tabular}{p{0.18\linewidth} p{0.72\linewidth}}
        \toprule
        \textbf{Symbol} & \textbf{Description} \\ 
        \midrule
        
        $r, \hat{r}$ & Factual and counterfactual dummy roots.\\
        $\mathcal{F}, \hat{\mathcal{F}}$ & Factual and counterfactual news feeds.\\
        $\mathcal{V}, \hat{\mathcal{V}}$ & Factual and counterfactual set of vertices.\\
        $\mathcal{E}, \hat{\mathcal{E}}$ & Factual and counterfactual set of edges.\\
        $\mathcal{U}$ & User profiles\\
        $u_j$ & Profile module of the $j$-th agent.\\
        $s_j, \hat{s}_j$ & Factual and counterfactual sensory modules of the $j$-th agent.\\
        $m_j, \hat{m}_j$ & Factual and counterfactual memory modules of the $j$-th agent.\\
        $b_j$ & Ban status of the $j$-th agent.\\
        $c_j$ & Counter of content violations the $j$-th agent.\\
        $a$ & Selected action.\\
        $p, \hat{p}$ & Factual and counterfactual parent nodes.\\
        $\mathcal{T}$ & Set of discussion topics.\\
        $x_{\mathit{user}}$ & Selected agent prompt template.\\
        $x_{\mathit{post}}$ & Post prompt template.\\
        $x_{\mathit{comm}}$ & Comment prompt template.\\
        $\psi$ & Prompting function.\\
        $o_j, \hat{o}_j$ & Factual and counterfactual post or comment by the $j$-th agent.\\
        $f$ & Toxicity detector.\\
        $\mathit{THR}$ & Moderation threshold.\\
        $\mathit{BAN}$ & Flag for BAN.\\
        $\mathit{OSFA}$ & Flag for One-Size-Fits-All.\\
        $\mathit{PMI}$ & Flag for Personalized-Moderation-Intervention.\\
        $x_{\mathit{mod}}$ & Moderator prompt template.\\
        \bottomrule
    \end{tabular}
\end{table}

\begin{figure*}[b]
    \centering
    \begin{tcolorbox}[colframe=gray, colback=white!95!white, boxrule=1.5pt]
    \small
    \textbf{Comment with memory (populated)}. \textit{You are now role-playing as a social network user. Below is your personal information:
    \\
    \\
    $<$personal information$>$\\
    Username: user\_21\\
    Age: 27\\
    Gender: female\\
    Race: white\\
    Income: high\\
    Education: high school\\
    Sex orientation: heterosexual\\
    Political leaning: independent\\
    Agreeableness: very low\\
    Openness: high\\
    Conscientiousness: low\\
    Extraversion: very low\\
    Neuroticism: high
    \\
    $<$/personal information$>$
    \\
    \\
    In the past, you have been moderated with the
    following intervention:
    \\
    \\
    $<$intervention$>$
    \\
    Dear user, your recent post violates our community guidelines due to the presence of offensive language. Please avoid this behavior in the future to help maintain a respectful community. Thank you for your cooperation.
    \\
    $<$/intervention$>$
    \\
    \\
    You are now reading the following thread:
    \\
    \\
    $<$thread$>$
    \\
    Post by user\_29: “As a very opinionated and extroverted foodie, I recently tried those overpriced vegan cupcakes from the latest trendy bakery in town. They're a disaster - dry, flavorless, and an insult to all things sweet. Those vegan hipsters need to go back to the drawing board.”
    \\
    $<$/thread$>$
    \\
    \\
    Step-by-step instructions:
    \\
    \begin{itemize}
        \item Suppose you want to write a comment on the thread, adopting the perspective of the social network user you are role-playing.
        \item Decide if you want to use toxic language (e.g. swears, insults, obscenities, identity attacks) based on your personality, past moderation and the context of the thread.
        \item Write the comment (up to 100 words).
        \item Enclose the comment within the tags $<$comment$>$ and $<$/comment$>$.
    \end{itemize}}
    \end{tcolorbox}
    \caption{Example of prompt template (comment with memory) filled with input information.}
    \label{fig:prompt_example}
\end{figure*}

\end{document}